\documentclass[twocolumn]{svjour3}          
\smartqed  
\usepackage{graphicx}
\usepackage{natbib}
\setcitestyle{round}

\usepackage{hyperref}       
\usepackage{url}            
\usepackage{booktabs}       
\usepackage{amsfonts}       
\usepackage{nicefrac}       
\usepackage{microtype}      

\usepackage{array}
\usepackage{bm}
\usepackage{wrapfig}
\usepackage{multirow}
\usepackage{threeparttable}

\usepackage{amsthm}

\usepackage{amssymb}
\usepackage{amsmath}
\usepackage{caption}

\usepackage{xcolor}
\newcommand{\sen}[1]{{\color{black} #1}}
\newcommand{\iclr}[1]{{\color{black} #1}}
\newcommand{\red}[1]{{\color{black} #1}}

\usepackage[misc]{ifsym} 

\begin{document}

\title{Information-Theoretic Odometry Learning
}


\author{Sen Zhang \and
        Jing Zhang \and
        Dacheng Tao 
}



\institute{Sen Zhang \at
              The University of Sydney, Australia. \\
              \email{szha2609@uni.sydney.edu.au}           
           \and
           \Letter~Jing Zhang \at
              The University of Sydney, Australia. \\
              \email{jing.zhang1@sydney.edu.au}   
           \and
           Dacheng Tao \at
              The University of Sydney, Australia. \\
              JD Explore Academy, China. \\
              \email{dacheng.tao@gmail.com}    
}


\date{Received: date / Accepted: date}

\maketitle

\begin{abstract}
In this paper, we propose a unified information theoretic framework for learning-motivated methods aimed at odometry estimation, a crucial component of many robotics and vision tasks such as navigation and virtual reality where relative camera poses are required in real time. We formulate this problem as optimizing a variational information bottleneck objective function, which eliminates pose-irrelevant information from the latent representation. The proposed framework provides an elegant tool for performance evaluation and understanding in information-theoretic language. Specifically, we bound the generalization errors of the deep information bottleneck framework and the predictability of the latent representation. These provide not only a performance guarantee but also practical guidance for model design, sample collection, and sensor selection. Furthermore, the stochastic latent representation provides a natural uncertainty measure without the needs for extra structures or computations. Experiments on two well-known odometry datasets demonstrate the effectiveness of our method.
\keywords{Odometry Learning \and Simultaneous Localization and Mapping \and Information Bottleneck \and Generalization Bound}
\end{abstract}

\section{Introduction}
\label{intro}
Odometry aims to predict six degrees of freedom (6-DOF) \sen{relative camera} poses from motion sensors. It is a fundamental component of a wide variety of robotics and vision tasks, including simultaneous localization and mapping (SLAM), automatic navigation, and virtual reality~\citep{durrant-whyte2006simultaneous, fuentes-pacheco2015visual, taketomi2017visual,zhang2020empowering}. In particular, visual and visual-inertial odometry have attracted a lot of attention over recent years due to the low cost and easy setup of cameras and inertial measurement unit (IMU) sensors. 
\sen{The relative camera pose is recovered using geometric clues and motion models. Classic geometric methods usually formulate the odometry problem as an optimization problem by incorporating well-established geometric and motion constraints as the objective functions. Nevertheless, due to the complexity and diversity of real-world environments, the explicitly modeled constraints can hardly explain all aspects of the sensor data. Though successful in some real-world scenarios, geometric systems fail to work when the underlying assumptions behind the optimization objectives, such as static environments, discriminative visual features, noiseless observations and brightness constancy, are violated in the real world.} 
Furthermore, since odometry is essentially a time-series prediction problem, how to properly handle time dependency and environment dynamics presents further challenges. 
\sen{Classic geometric methods use filtering or bundle adjustments to take the temporal information into account, while the implicitly implied error distributions might not hold in practice.}


\sen{Recently end-to-end deep learning methods provide an alternative solution for the odometry problem, which relieves the above-mentioned intrinsic problems in geometric methods. Learning-based methods tackle this problem from another perspective that does not explicitly model the constraints for optimization but learns the mapping from sensor data to camera pose implicitly from large-scale datasets~\citep{wang2017deepvo, clark2017vinet, xue2019beyond}. It has been shown that well-trained deep networks are able to effectively capture the inherent complexity and diversity of the training data and establish the mapping between visual/sequential inputs to desired targets in many computer vision tasks, thus holding promise for addressing the limitations of geometric approaches. In addition, learning-based frameworks can implicitly learn calibrated representations and require no explicit calibration procedures. For monocular visual odometry, the absolute scale can also be recovered from training data, which instead is a non-trivial challenge for geometric methods.}

Although existing deep odometry learning methods have performed competitively against their \iclr{geometric counterparts}, they still fail to satisfy some basic requirements. First of all, due to the broad range of scenarios where odometry is required, odometry systems are expected to be easily compatible with various configurations and settings, such as multiple sensors and dynamic environments. \sen{In addition}, the common existence of data degeneration, such as from hardware malfunctions and unexpected occlusions, requires a safe and robust system in which a proper uncertainty measure is desirable for self-awareness of the potential anomalies and system bias. Moreover, theoretical analyses of current black box deep odometry models, such as generalizability on unseen test data and extendibility to extra sensors, are still obscure but essential for understanding and assessing the model performance.

Here we devise a unified odometry learning framework from an information-theoretic perspective, which well addresses the above issues. Our work is motivated by the recent successes of deep variational inference and learning theory based on mutual information (MI). Specifically, we translate the odometry problem to optimizing an information bottleneck (IB) objective function where the latent representation is formulated as a bottleneck between the observations and relative camera poses. In doing so, we eliminate the pose-irrelevant information from the latent representation to achieve better generalizability. Modeling by MI constraints provides a flexible way to account for different aspects of the problem and quantify their effectiveness in information-theoretic language. This framework is also attractive in that the operations are performed on the probabilistic distribution of the latent representation, which naturally provides an uncertainty measure for interrogating the data quality and system bias.

More importantly, the information-theoretic formulation allows us to leverage information theory to investigate the theoretical properties of the proposed method. Our theoretical findings not only benefit the evaluation of the model performance but also provide insights for subsequent research. We obtain a theoretical guarantee of the proposed framework by deriving an upper bound of the expected generalization error w.r.t. the IB objective function under mild network and loss function conditions. We show that the latent space dimensionality also bounds the expected generalization error, providing a theoretical explanation for the complexity-overfitting trade-off in the latent representation space. When the test data is biased, our result shows that the growing rate of $d$ should not exceed that of  $n/log(n)$, where $d$ is the latent space dimensionality, and $n$ is the sample size. We further quantify the usefulness of a latent representation for relative camera pose prediction using the MI between the representation and poses. In doing so, we prove a lower bound for this MI given extra sensors, which reveals the conditions required for a sensor to theoretically guarantee a performance gain. It is noteworthy that our theoretical results hold not only for the odometry problem but also for a wider variety of problems that share the same Markov chain assumption and the IB objective function. \iclr{A connection between our information-theoretic framework and geometric methods is further established for deeper insights.} 

The main contributions of this paper are:
\begin{enumerate}
    \item We propose information-theoretic odometry learning by leveraging the IB objective function to eliminate pose-irrelevant information from the latent representation; 
    
    \item We develop the theoretical performance guarantee of the proposed framework by deriving upper bounds on the generalization error w.r.t. IB and the latent space dimensionality as well as a lower bound on the MI between the latent representation and poses; 
    
    \item We empirically verify the effectiveness of our method on the well-known KITTI and EuRoC datasets and show how the intrinsic uncertainty benefits failure detection and inference refinement. 
\end{enumerate} 

\section{Related Work}

\paragraph{Deep representation for odometry learning:} Leveraging deep neural networks to learn compact feature representation from high-dimension sensor data has been proven effective for odometry. \cite{kendall2015posenet} proposed PoseNet by using neural networks for camera relocalization, based upon which \cite{wang2017deepvo} introduced a recurrent module to model the temporal correlation of features for visual odometry. Subsequently, \cite{xue2019beyond} further considered a memory and refinement module to address the prediction drift caused by error accumulation. Recently, deep learning-based odometry has also been extended to the multi-sensor configuration. \cite{clark2017vinet} extended the DeepVO framework to incorporate IMU data by leveraging an extra recurrent network for learning better feature representation. A recent study by \cite{chen2019selective} investigated more effective and robust sensor fusion via soft and hard attention for visual-inertial odometry. \iclr{Apart from end-to-end learning, there are also trends in unsupervised learning~\citep{zhou2017unsupervised,yin2018geonet,ranjan2019competitive,bian2019unsupervised} and the combination of learned features with geometric methods~\citep{zhan2019visual,yang2020d3vo}. We refer readers to \citet{chen2020a} for a more detailed discussion of current methods.} These deep odometry learning methods have achieved promising performance. However, theoretical understandings remain obscure: (1) how to learn a compact representation with a theoretically guaranteed generalizability when test data is biased and (2) in what conditions extra sensors can benefit the pose prediction problem. 

\begin{figure*}
    \centering
    \includegraphics[width=.9\textwidth]{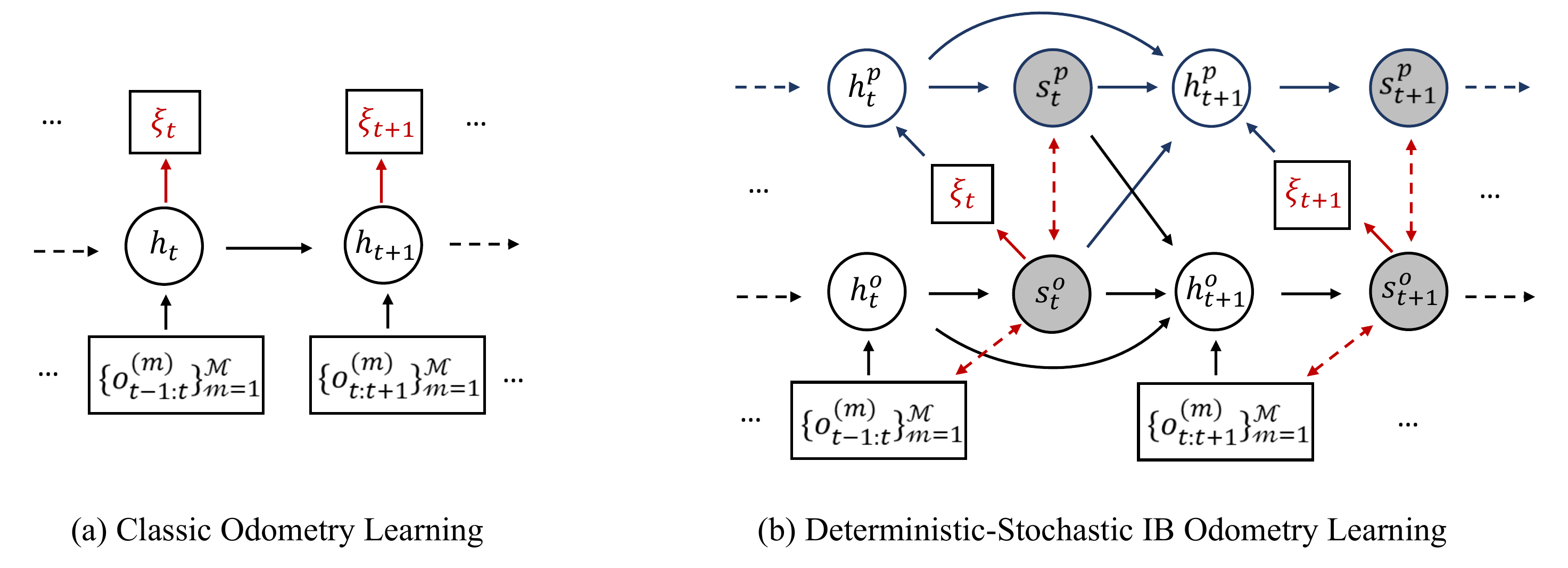}
    \caption{(a) The classic learning-based odometry framework, where 6-DOF poses are directly predicted from deterministic latent representations. (b) The proposed information bottleneck (IB) framework for odometry learning. $h$ and $s$ are the deterministic and stochastic components, respectively. Superscripts $o$ and $p$ represent the observation- and pose-level transition models. Red solid arrows denote the pose regressor, and red dashed arrows denote the bottleneck constraints. Output arrows from a shaded stochastic representation represent samples from the learned latent distribution.}
    \label{fig:framework_1}
\end{figure*}

\paragraph{Information bottleneck:} 
Information bottleneck (IB) provides an appealing tool for deep learning by learning an informative and compact latent representation~\citep{tishby2000the, tishby2015deep, shwartz-ziv2017opening}. To address the intractability of MI calculation, \cite{alemi2017deep} proposed to optimize a variational bound of IB for deep learning, which was successfully applied to many tasks including dynamics learning~\citep{hafner2020dream}, task transfer~\citep{goyal2019infobot}, and network compression~\citep{dai2018compressing}. Partly inspired by these developments, we for the first time propose an IB-based framework for odometry learning \iclr{and derive an optimizable variational bound for this sequential prediction problem. The derivation can be more delicate if we incorporate more constraints, potentially from geometric and kinematic insights. We further adopt the deterministic-stochastic separation as in \citet{chung2015a,hafner2019learning,hafner2020dream}, while ours differs in that our derivation of the variational bound allows modeling two transition models separately, each with a deterministic component to improve model capacity.  Moreover,} though IB-based methods have shown to be effective for learning a compact representation, the underpinning generalizability theory remains unclear. The generalization error bounds for general learning algorithms have been studied in \cite{xu2017information} in information-theoretic language. This work was subsequently extended by \cite{zhang2018an} to explain the generalizability of deep neural networks. However, their results are not applicable to the IB-based methods, which will be addressed in this paper.

\paragraph{Uncertainty modeling for odometry learning:}
Modeling uncertainty to deal with extreme cases like hardware malfunctions and unexpected occlusions, is crucial for a reliable and robust odometry system. It can be categorized into model-intrinsic epistemic uncertainty and data-dependent aleatoric uncertainty, which have been studied in the Bayesian deep learning literature~\citep{mackay1992a, gal2016dropout, kendall2017what}. For odometry, \cite{wang2018end} and \cite{yang2020d3vo} captured the aleatoric uncertainty by imposing a probabilistic distribution on poses and used the second moment prediction as an uncertainty measure. Recently, \cite{loquercio2020a} showed that a combined epistemic-aleatoric uncertainty framework~\citep{kendall2017what} could improve the performance on several robotics tasks such as motion and steering angle predictions. In contrast to them, our framework provides a built-in and efficient uncertainty measure that accounts for both uncertainty types. We empirically demonstrate how to use this uncertainty measure to evaluate data quality and system biases. Accordingly, we propose a refined inference procedure that discards highly uncertain results to improve pose prediction accuracy.

\section{Information-Theoretic Odometry Learning}

Odometry aims to predict the relative 6-DOF pose $\xi_t$ between two consecutive observations $\{o_{t-1:t}^{(m)}\}_{m=1}^{\mathcal{M}}$ from $\mathcal{M}$ sensors (e.g. camera, IMU and lidar), where $t$ is the time index. This pose prediction problem can be formulated as $\xi_t = g(\{o_{t-1:t}^{(m)}\}_{m=1}^{\mathcal{M}}, \Theta)$, where $g$ is the mapping function of an odometry system and $\Theta$ is the parameter set of $g$. Classic deep odometry learning methods model $g$ by neural networks and learn $\Theta$ from training data. Furthermore, they usually use a recurrent module to model the motion dynamics of the observation sequence. Fig.~\ref{fig:framework_1}(a) shows a typical procedure shared by representative deep odometry learning methods.

In many settings, observations are of high dimensionalities, such as images and lidar 3D points. \iclr{Geometric methods use low-dimensional features} to represent observations, while learning-based methods learn a representation from training data. However, both features may contain pose-irrelevant information that is specific to certain sensor domain. Retaining such information encourages the model to overfit the training data and yield poor generalization performance. Since parsimony is preferred in machine learning, it is expected to eliminate the pose-irrelevant information. 

To this end, we tackle this problem by explicitly introducing a constraint on the pose-irrelevant information. Specifically, we quantify the pose-irrelevance and the usefulness of a latent representation for pose prediction from an information-theoretic perspective. By assuming the latent representation $s_t$ at time $t$ is drawn from a Gaussian distribution, the MI $I(\{o_{1:T}^{(m)}\}_{m=1}^{\mathcal{M}}||s_{1:T}|\xi_{1:T})$ and the MI $I(\xi_{1:T} || s_{1:T})$ can provide quantitative measures for the aforementioned two aspects. Accordingly, given a sequence of observations $\{o_{1:T}^{(m)}\}_{m=1}^{\mathcal{M}}$ and pose annotations $\xi_{1:T}$ from time $1$ to $T$, our information-theoretic odometry learning problem is:
\begin{eqnarray}
    max_{\Theta}\ \mathcal{J}(\Theta) &=& I(\xi_{1:T} || s_{1:T}) - \gamma I_{bottleneck},
    \label{eq:ib_objective} \\
    I_{bottleneck} &=& I(\{o_{1:T}^{(m)}\}_{m=1}^{\mathcal{M}}||s_{1:T}|\xi_{1:T}),
\end{eqnarray}
where the IB weight $\gamma$ controls the trade-off between the two MI terms. By Equation~\ref{eq:ib_objective}, the latent representation $s_{1:T}$ essentially provides an information bottleneck between poses and observations, which eliminates pose-irrelevant information from the observations. Due to the high dimensionality of the observation space, it is non-trivial to calculate the two MI. \iclr{Thus we optimize a variational lower bound instead:}
\begin{eqnarray}
    \mathcal{J}(\Theta)\geq \mathcal{J}'(\Theta)&=& E_{s_{1:T},\{o_{1:T}^{(m)}\}_{m=1}^{\mathcal{M}},\xi_{1:T}} [\sum\nolimits_{t=1}^T J'_t], \label{eq:obj_variational}\\
    J'_t &=& \mathcal{J}_t^{pose} - \gamma \mathcal{J}_t^{bottleneck}, \label{eq:bottleneck_weight}\\
    \mathcal{J}_t^{pose} &=& log\ q_\theta(\xi_t|s_t),  \label{eq:obj_pose} \\
    \mathcal{J}_t^{bottleneck} &=& D_{KL}(p_\phi || q_\varphi). \label{eq:obj_kl}, \\
    p_\phi &=& p_\phi(s_t|\{o_{t-1:t}^{(m)}\}_{m=1}^{\mathcal{M}},s_{t-1}), \\
    q_\varphi &=& q_\varphi(s_t|\xi_t, s_{t-1}).
\end{eqnarray}
The detailed derivation is provided in the Supplementary Material. This lower bound consists of a variational pose regressor $q_\theta(\xi_t|s_t)$, an observation-level transition model $p_\phi(s_t|\{o_{t-1:t}^{(m)}\}_{m=1}^{\mathcal{M}},s_{t-1})$, and a pose-level transition model $q_\varphi(s_t|\xi_t, s_{t-1})$, all of which are modeled by neural networks. 
For simplicity, we denote the representations from the observation-level and pose-level transition models $s^o_t$ and $s^p_t$, respectively. In practice, $s^o_t$ is used for the pose regressor. Intuitively, minimizing the KL divergence in Equation~\ref{eq:obj_kl} forces the distribution of $s_t^o$ to approximate that of $s_t^p$ which does not encode the observation information at time $t$, thus regularizing $s_t^o$ for containing pose-irrelevant information. 

Stochastic-only transition models, however, may compromise model performance due to uncertainty accumulation during the sampling process. To address this problem, we further introduce a deterministic component according to \cite{chung2015a} and \cite{hafner2019learning}. In doing so, we reformulate the two transition models in the KL divergence in Equation~\ref{eq:obj_kl} as:
\begin{eqnarray}
    &\textrm{\textbf{observation-level}}:\  p_\phi(s_t^o|h_t^o), \label{eq:trans_obs}\\ 
    &h_t^o = f^o(h_{t-1}^o,\{o_{t-1:t}^{(m)}\}_{m=1}^{\mathcal{M}}, s_{t-1}^o, s_{t-1}^p), \\
    &\textrm{\textbf{pose-level}}:\  q_\varphi(s_t^p|h_t^p), \label{eq:trans_pose} \\ 
    &h_t^p = f^p(h_{t-1}^p,\xi_t, s_{t-1}^o, s_{t-1}^p).
\end{eqnarray}

We use two deterministic functions $f^o$ and $f^p$ for observation- and pose-level transitions, respectively, which are both modeled by recurrent neural networks. In addition, both $s^o_{t-1}$ and $s^p_{t-1}$ are used for the two deterministic transition functions to help to reduce the KL divergence between the distributions of $s_t^o$ and $s_t^p$. Ground-truth 6-DOF poses are fed into $f^p$ during the training phase, while for testing, we use predicted poses to provide a runtime estimate of $s_t^p$. Fig.~\ref{fig:framework_1}(b) shows the overall framework of our method.

\textbf{Remark I:}
Since we model the latent representation in the probabilistic space, the variance of the latent representation naturally provides an uncertainty measure. We empirically show how this intrinsic uncertainty reveals data quality and system bias in Section 5.3. Of note is that it is straightforward to extend the proposed information-theoretic framework to different problem settings. We can add arbitrary linear MI constraints into the proposed objective \iclr{and derive similar variational bounds} to satisfy different requirements such as dynamics-awareness in complex environments. 

\textbf{Remark II:}
All variational IB-based methods origin from \citet{alemi2017deep}. However, applying IB into a specific domain is non-trivial. The challenge lies in the derivation of proper variational bounds based on the specific properties of each problem. This derivation can be more delicate if we incorporate more constraints, potentially from geometric and kinematic insights. Besides, we differ from \citet{dai2018compressing} and \citet{goyal2019infobot} in that sequential observations are modeled. From this perspective, our development related to \citet{hafner2019learning} and \citet{hafner2020dream}, from which we further borrowed the motivation of the deterministic component, which by itself is rooted from \citet{chung2015a} and \citet{buesing2018learning}. Ours differs in that we model the two transition models (Equation~\ref{eq:obj_kl}) separately, each with a deterministic component to improve model capacity (Fig.~\ref{fig:framework_1}(b) and Equations~\ref{eq:trans_obs}-\ref{eq:trans_pose}). Moreover, we theoretically prove that constraining the IB objective essentially upper bounds the expected generalization error and establish the connection between IB and geometric methods in Section 4, which provides deeper insights into IB-based methods.

\section{Theoretical Analysis}

Formulating a problem in information-theoretic language enables us to analyze the proposed method by exploring elegant tools in information theory~\citep{cover1991elements} and related results in learning theory~\citep{xu2017information, zhang2018an}. In this work, we show that the MI between the bottleneck and observations as well as the latent space dimensionality upper bound the expected generalization error, which provides not only insights into the generalizability of the method but also a performance guarantee. To our knowledge, this is the first time that such generalization bounds have been derived for IB by using a general loss function other than cross-entropy~\citep{vera2018the}. By replacing the general loss function with the cross-entropy, our bound is tighter than that obtained by \cite{vera2018the} in terms of the sample size. We further derive a lower bound on the MI between the latent representation and poses given extra sensors, which suggests what features make a sensor useful for pose prediction in information-theoretic language. The connection between information bottleneck and geometric methods is also established to provide further insights.

\subsection{Generalization Bound for Information Bottleneck}

\cite{xu2017information} and \cite{zhang2018an} obtained the generalization bound w.r.t. the MI between input data $X$ and learning parameters $\Theta$ for general learning algorithms and neural networks. However, what IB regularizes is the MI between $X$ and the latent representation. To derive a generalization bound for the IB objective function, we first prove a relationship between these two kinds of MI in Lemma~\ref{lemma:1} under the Markov chain $X\rightarrow S\rightarrow \xi$, an underlying assumption for IB.
\begin{lemma}
    If $X\rightarrow S\rightarrow \xi$ forms a Markov chain and assume $\xi=g(X,\Theta)$ is a one-to-one function w.r.t. $X$ and $\Theta$, then we have
    \begin{equation}
        I(X,S) \geq I(X,\xi) = I(X, \Theta) + E_{\theta}[H(X|\theta)] 
    \end{equation}
    \begin{equation}
        \ \ \ \ \ \ \ \ \ \ \geq I(X,\Theta).
    \end{equation}
    \label{lemma:1}
\end{lemma}
Lemma~\ref{lemma:1} enables us to extend the generalizability results for neural networks regarding $I(X,\Theta)$~\citep{zhang2018an} to the IB setting, leading to the following theoretical counterpart:
\begin{theorem}
   Assuming $X\rightarrow S\rightarrow \xi$ is a Markov chain, the loss function $l(X,\Theta)$ is sub-$\sigma$-Gaussian distributed\footnote{Recall that a random variable $l$ is sub-$\sigma$-Gaussian distributed if $E[e^{\lambda(l-E[l])}]\leq e^{\frac{\lambda^2\sigma^2}{2}},\ \forall\  \lambda\in R$.} and the prediction function $\xi=g(X,\Theta)$ is a one-to-one function w.r.t. the input data and network parameters $\Theta$, we have the following upper bound for the expected generalization error:
    \begin{equation}
        E[R(\Theta)-R_T(\Theta)]\leq exp(-\frac{L}{2}log\frac{1}{\eta})\sqrt{\frac{2\sigma^2}{n}I(X,S)},
    \end{equation}
    where $L$, $\eta$, and $n$ are the effective number of layers causing information loss, a constant smaller than 1, and the sample size, respectively. $R(\Theta)=E_{X\sim D}[l(X,\Theta)]$ is the expected loss value given $\Theta$ and $R_T(\Theta)=\frac{1}{n}\sum_{i=1}^n l(X_i,\Theta)$ is a sample estimate of $R(\Theta)$ from the training data.
    \label{theorem:1}
\end{theorem}

The difference between our result and previous works is that we bound the generalization error by $I(X,S)$ which is minimized in Equation~\ref{eq:ib_objective} rather than $I(X,\Theta)$ which is hard to evaluate. By Theorem~\ref{theorem:1}, we show that minimizing the MI between the bottleneck and observations tightens the upper bound on the expected generalization error and thus provides a theoretical performance guarantee. It is worth noting that our theoretical results apply not only to our odometry learning setting but also to a wider variety of tasks that use the IB method. This bound also implies that a larger sample size and a deeper network lead to better generalization performance, which is consistent with the results shown in \cite{xu2017information} and \cite{zhang2018an}. The detailed proof of Lemma~\ref{lemma:1} and Theorem~\ref{theorem:1} can be found in the Supplementary Material. 

\textbf{Remark I:} 
The result of \cite{zhang2018an} is interesting in that it provides an explanation for why deeper networks lead to better performance. However, the expected generalization errors in \cite{zhang2018an} and \cite{xu2017information} are both bounded by $I(X||\Theta)$, which remains difficult to evaluate in practice. Though their results give a lot of insights into the generalizability of algorithms in information-theoretic language, it is non-trivial to minimize $I(X||\Theta)$ explicitly to control the generalization error bound. 
We move one step further by extending their results to $I(X||S)$, the mutual information between input data and latent representations, which itself can be bounded by various well-established variational bounds~\citep{poole2019on} and optimized during training. Our result provides an explanation for the empirical generalization ability of the IB method, which explicitly minimizes $I(X||S)$. By minimizing $I(X||S)$, we are actually tightening the upper bound of the generalization error, thus leading to better generalization performance. 

A related work by \cite{vera2018the} proved a similar result for IB: "Let $\mathcal{F}$ be a class of encoders. Then, for every $P_{XY}$ and every $\delta\in(0,1)$, with probability at least $1-\delta$ over the choice of $\mathcal{S}_n\sim P_{XY}^n$ the following inequality holds $\forall Q_{U|X}\in\mathcal{F}$:
\begin{eqnarray}
    \varepsilon_{gap}(Q_{U|X}, \mathcal{S}_n)\leq A_\delta\sqrt{I(\hat{P_X}||Q_{U|X})}\frac{log(n)}{\sqrt{n}} \nonumber 
    \\+\frac{C_\delta}{\sqrt{n}}+\mathcal{O}(\frac{log(n)}{n}), \label{vera}
\end{eqnarray}
where $(A_\delta, B_\delta, C_\delta)$ are quantities independent of the data set $\mathcal{S}_n: A_\delta:=\frac{\sqrt{2}B_\delta}{P_X(x_{min})}(1+1/\sqrt{|X|}),B_\delta:=2+\sqrt{log(\frac{|Y|+3}{\delta})}$ and $C_\delta:=2|U|e^{-1}+B_\delta\sqrt{|Y|}log\frac{|U|}{P_Y(y_{min})}$. $\varepsilon_{gap}(Q_{U|X}, S_n)$ is the generalization gap which is defined as $|L_{emp}(Q_{U|X}, \mathcal{S}_n)--L(Q_{U|X})|$. $L(Q_{U|X})$ and $L_{emp}(Q_{U|X}, \mathcal{S}_n)$ are the true risk and the empirical risks, respectively." We refer readers to \cite{vera2018the} for more details on their result 

Our result differs from that of \cite{vera2018the} in that: (1) Equation~\ref{vera} only applies to the cross-entropy loss function, while our result holds for a broader range of loss functions under the sub-$\sigma$-Gaussian assumption; (2) We provide a tighter generalization bound compared with that of \cite{vera2018the} w.r.t. sample rate ($\frac{1}{\sqrt{n}}$ vs. $\frac{log(n)}{\sqrt{n}}$); (3) For regression problems and for a large latent space, $A_\delta$ and $C_\delta$ in Equation~\ref{vera} could be large due to the positive dependency on $|Y|$ and $|U|$. Besides, $\frac{1}{P_X(x_{min})}$ and $\frac{1}{P_Y(y_{min})}$ might also be large in practice, resulting in a loose bound for the generalization error.

\textbf{Remark II:}
We now give more discussions on the assumptions of Theorem~\ref{theorem:1}: (1) A Markov chain $X\rightarrow S\rightarrow \xi$ is implicitly implied in neural networks with encoder-decoder structures since the decoder only takes the encoder output as its input and thus does not depend on $X$ given $S$. Since an IB model is essentially encoder-decoder structured by constraining the information flow between the encoder and the decoder, the Markov chain assumption on $X\rightarrow S\rightarrow \xi$ also holds for the IB methods. (2) As discussed in \cite{xu2017information}, the sub-$\sigma$-Gaussian assumption actually implies a broad range of loss functions. For instance, as long as a loss function $l$ is bounded, i.e., $l(\cdot,\cdot)\in[a,b]$, then it is guaranteed to be sub-$\sigma$-Gaussian distributed with $\sigma=\frac{b-a}{2}$~\citep{xu2017information}. \iclr{The network loss landscape consists of multiple local minima, flat or sharp, and most deep learning methods assume a local Gaussian distribution by using L2 loss~\citep{chaudhari2017entropy}. Sub-$\sigma$-Gaussian is more general and provides several superiorities over the commonly used Gaussian assumption. \citet{chaudhari2017entropy} claimed that a flat local minimum is preferred for deep learning optimization algorithms due to the robustness towards parameter perturbations. Sub-$\sigma$-Gaussian can well represent such flat local regions, e.g. the almost-flat bounded uniform distribution is sub-$\sigma$-Gaussian distributed. It is also worth noting that considering the density of local minima~\citep{chaudhari2017entropy}, $\sigma$ is not necessarily large for local regions, which can be a concern for the tightness of the generalization bound. Another appealing property is that the sum of sub-$\sigma$-Gaussian is still sub-$\sigma$-Gaussian, i.e. it can fit a larger region with multiple local minima.} (3) The one-to-one function assumption can be conservative due to the complexity of real-world data. For many applications, we may use pretrained models to extract high-level features and use these features as input data. For example, a pretrained FlowNet~\citep{dosovitskiy2015flownet,ilg2017flownet} is usually used in deep odometry learning methods. The input data part of this assumption could arguably hold under such circumstances. Considering the prediction part of this assumption, the cardinality of the space of $\xi$ could be sufficiently large for regression problems and for classification problems, the cardinality of the prediction space could also be large since we usually predict the probabilities of each category. Extending the results to a looser assumption on the network function remains an interesting direction for future research.

\subsection{Generalization Bound for Latent Dimensionality}

We further investigate the generalizability w.r.t. model complexity in terms of the cardinality and dimensionality of the latent representation space under the IB framework. 
\begin{corollary}
     Given the same assumptions in Theorem~\ref{theorem:1} and let |S| be the cardinality of the latent representation space, we have
    \begin{equation}
        E[R(\Theta)-R_T(\Theta)]\leq exp(-\frac{L}{2}log\frac{1}{\eta}) \sqrt{\frac{2\sigma^2}{n}log|S|}.
    \end{equation}
    \label{corollary:1}
\end{corollary}
It is well recognized that a large model complexity can impair the generalizability of the model. We reveal this complexity-overfitting trade-off in Corollary~\ref{corollary:1}, where the expected generalization error is upper bounded by the cardinality of the latent representation space. In addition, considering the model design and sample collection, Corollary~\ref{corollary:1} indicates that the growing rate of $log|S|$ should not exceed that of  $n$ to avoid an exploded generalization error bound. 

\begin{corollary}
  Given the same assumptions in Theorem~\ref{theorem:1} and assume $S$ lies in a d-dimensional subspace of the latent representation space, $sup_{s_i\in S_i}\ ||s_i|| \leq M,\forall i\in [1,d]$ and S can be approximated by a densely quantized space, the following generalization bound holds:
   \begin{equation}
    E[R(\Theta)-R_T(\Theta)]\leq exp(-\frac{L}{2}log\frac{1}{\eta})\sigma \mathcal{C}, 
   \end{equation}
   \begin{equation}
    \mathcal{C} = \sqrt{\frac{dlog (d)}{n}+2log(2M)\frac{d}{n}+\frac{d}{n/log(n)}}.
   \end{equation}
   \label{corollary:2}
\end{corollary}

In practice, it is usually difficult to evaluate $log|S|$ in Corollary~\ref{corollary:1} numerically. Therefore, we leverage the quantization trick used in \cite{xu2017information} to reduce the upper bound to a function w.r.t. the dimensionality $d$ of the latent representation space. The result is given in Corollary~\ref{corollary:2}, which suggests that the growing rate of $d$ should not exceed that of $n/log(n)$. It is worth noting that this result holds not only for IB but also for a broader range of encoder-decoder models under the Markov chain assumption on $X\rightarrow S\rightarrow \xi$. 

\subsection{Predictability Bound for Extra Sensors}

Odometry performance is highly dependent on the sensors deployed, yet it remains non-trivial to select informative sensors that guarantee a performance gain. In this section, we address this problem using information-theoretic language under our proposed framework.

\begin{theorem}
    If $(\{o^{(m)}\}_{m=1}^{\mathcal{M}},\ o^{(\mathcal{M}+1)})\rightarrow S\rightarrow \xi$ forms a Markov chain, then we have,
    \begin{equation}
        I(\xi||S) \geq I_{old} + I_{new} - I_{obs},
    \end{equation}
    \begin{equation}
        I_{old} = I(\xi||\{o^{(m)}\}_{m=1}^{(\mathcal{M})}),
    \end{equation}
    \begin{equation}
         I_{new} = I(\xi|| o^{(\mathcal{M}+1)} | \{o^{(m)}\}_{m=1}^{\mathcal{M}}),
    \end{equation}
    \begin{equation}
        I_{obs} = I(o^{(\mathcal{M}+1)}||\{o^{(m)}\}_{m=1}^{\mathcal{M}}|\xi).
    \end{equation}
    \label{theorem:2}
\end{theorem}
Theorem~\ref{theorem:2} suggests that if a new sensor $o^{(\mathcal{M}+1)}$  is useful for pose prediction, the MI between $o^{(\mathcal{M}+1)}$ and poses given existing sensors should be large. Meanwhile, it is preferred to have a small MI between $\{o^{(m)}\}_{m=1}^{(\mathcal{M})}$ and $o^{(\mathcal{M}+1)}$ given pose information. In other words, a heterogenous sensor that shares little pose-irrelevant information with existing sensors is desirable. In addition, we further observe that the information gain between $I(\xi|| o^{(\mathcal{M}+1)} | \{o^{(m)}\}_{m=1}^{\mathcal{M}})$ and $I(o^{(\mathcal{M}+1)}||\{o^{(m)}\}_{m=1}^{\mathcal{M}}|\xi)$ provides a theoretical guarantee for the performance of the learned latent representation.

\iclr{\subsection{Connection with Geometric Methods}
More generally, an odometry system can be modeled as $h(z_{k,j}, v_k, \check{x_k})\rightarrow (\hat{x_k},p_j)$ where $z_{k,j}, v_k, \check{x_k}, \hat{x_k}$ and $p_j$ are observations, noise, prior pose, posterior pose, and latent state, respectively. At this level, the bottleneck MI $I(z_{k,j},v_k||p_j|\hat{x_k})=H[h(z_{k,j},v_k,\check{x_k})|\hat{x_k}]-H[h(z_{k,j},v_k,\check{x_k})|\hat{x_k},z_{k,j},v_k]$ is the extra entropy ($\Delta H$) introduced by $(z_{k,j},v_k)$, which differs for different $h$. Factor graph based methods use optimization over L2 costs as $h$, where $p_j$ is inferred landmark and a Gaussian noise is assumed. $\Delta H$ in this case is implied in the noise variance which corresponds to the pre-specified weight of each cost function. Learning-based methods learn $h$ from data where $p_j$ is the latent feature. Minimizing $\Delta H$ means reducing the uncertainty from noise and inexact learned function forms. The same analysis applies to kinematic function for $\check{x_k}$. \sen{In addition}, filter-based methods can also be included in by following the same logic. Take the kinematics part of Kalman filter (linear Gaussian system) as an example: $\check{x_k}=A_k\hat{x_{k-1}}+u_k+w_k$, where the prior $\check{x_k}$ is the latent state and the variance of $\hat{x_{k-1}}$ and $w_k$ are $\hat{\Sigma_{k-1}}$ and $R$, respectively. Then $I(u_k,w_k||\check{x_k})=\frac{1}{2}ln(|A_k\hat{\Sigma_{k-1}}A_k^T+R|/|A_k\hat{\Sigma_{k-1}}A_k^T|)$, suggesting that a smaller bottleneck MI corresponds to a relatively smaller noise variance.}

\section{Experiments}
We tested our method on the well-known KITTI~\citep{geiger2013vision} and EuRoC~\citep{burri2016the} datasets. Since most existing supervised methods are not open source, we re-implemented the representative state-of-the-art methods, including DeepVO~\citep{wang2017deepvo}, VINet~\citep{clark2017vinet}, and two attention-based visual-inertial methods recently proposed by \cite{chen2019selective}, namely, SoftFusion and HardFusion, as our baselines. All models shared the same network architecture for a fair comparison. \red{We further examine the ability of generalization to more challenging scenarios such as extreme weather and lighting conditions by testing DeepVO and InfoVO on vKITTI2~\citep{cabon2020virtual}. In addition, we empirically study the pose-irrelevant information contained in DeepVO and InfoVO to examine the underlying hypothesis of the problem that we target.} We also conducted extensive ablation studies on the deterministic component, the weight of the IB objective, the sample size, extra sensors, the intrinsic uncertainty measure\red{, and the growing rate relationship between the latent dimension and $n/log(n)$}. 

\subsection{Datasets and Experimental Settings}
The KITTI odometry dataset consists of 11 real-world car driving videos and calibrated ground-truth 6-DOF pose annotations. The EuRoC dataset was instead collected from a MAV in two buildings, resulting in 11 sequences of different difficulties by manually adjusted obstacles. For visual-inertial experiments, we manually aligned the 100 Hz IMU records in the raw KITTI dataset to the 10 Hz image sequences using the corresponding timestamps. The image and IMU sequences in EuRoC were downsampled to 10 Hz and 100 Hz, respectively. We split the training and test datasets following the recent work by \cite{chen2019selective}. Our implementation was based on PyTorch~\citep{steiner2019pytorch}, and we will release the source code package and the trained models. We used GRU~\citep{cho2014learning} to model the deterministic transitions and IMU records. Pretrained FlowNet was used to extract features from image data~\citep{dosovitskiy2015flownet, ilg2017flownet}. The other parts were modeled by MLP layers. 

\subsubsection{Detailed Network Architecture}
The overall network can be separated into four components: \textbf{(1) Observation encoders}: For image observation, we first extract the output from the \emph{$out\_conv6\_1$} layer of a pretrained FlowNet2S~\citep{ilg2017flownet} model as an intermediate high-level feature, which is then flattened and fed into three MLP layers that have feature size 1024 to obtain image features. Note that the last MLP layer does not use the non-linear activation. For IMU data, we use a two-layer GRU model that has feature size 1024 to extract IMU features; \textbf{(2) Deterministic transition models}: For the observation-level transition, we first fuse the observation features and concatenate the fused feature with $s^o_{t-1}$ and $s^p_{t-1}$ from last time step. The features are concatenated in VINet and InfoVIO. For SoftFusion, SoftInfoVIO, HardFusion and HardInfoVIO, we also use the same soft and hard fusion strategy proposed in \cite{chen2019selective}, while the Gumbel temperature linearly degrades from 1 to 0.5 in the first 150 epochs during training and is fixed to 0.5 for testing. We tile the 6-DOF poses eight times to a vector of length 48 for the pose-level transition, which is then also concatenated with $s^o_{t-1}$ and $s^p_{t-1}$. Ground-truth 6-DOF poses are used during training, while the predicted poses are used during testing. The concatenated features are then fed into an MLP and a GRU layer to obtain $h^o_t$ and $h^p_t$, respectively. \textbf{(3) Stochastic state estimators}: The deterministic states are fed into two MLP layers to obtain the mean and standard error vectors of the stochastic representation, both with size 128. Note that the last MLP layer does not use the non-linear activation. To avoid a trivial solution, we set the minimum standard error to 0.1 and only predict the residue, where the softplus function is used to guarantee a positive residue. We further use the reparameterization trick proposed in \cite{kingma2014auto} to sample from the stochastic representation distributions, which enables gradient backpropagation through the stochastic representations. \textbf{(4) Pose regressor}: We feed the sampled observation-level representation $s^o_t$ into three MLP layers to obtain the translation and rotation prediction results. Both translation and rotation share the first two MLP layers, while we use two separate MLP layers without non-linear activation for translation and rotation, respectively. 

All MLP layers with non-linear activation use the Relu function and have feature sizes 256 and 512 for KITTI and EuRoC, respectively. The state size is set to 128 and 256 for KITTI and EuRoC, respectively. For all baseline models (DeepVO, VINet, SoftFusion, and HardFusion), we remove the pose-level transitions and stochastic state estimators and directly feed $h^o_t$ into the pose regressor for prediction. 

\subsubsection{Training and Evaluation Strategies}
We used the same training and test splits as \cite{chen2019selective}. For KITTI, we used sequences 00, 01, 02, 04, 06, 08, and 09 for training and the rest for testing. For EuRoC, we used the sequence \emph{MH\_04\_difficult} for testing and the rest for training. KITTI odometry dataset does not contain synchronized IMU data. Therefore, we manually aligned the 100 Hz IMU records in the raw KITTI data to the 10 Hz image sequences using the corresponding timestamps. EuRoC provides synchronized image and IMU data, collected at 20 Hz and 200 Hz, respectively. Following the practice of previous work~\citep{chen2019selective, clark2017vinet}, we downsampled the image and IMU data in EuRoC to 10 Hz and 100 Hz, respectively. By assuming a Gaussian distribution for $q_\theta(\xi_t|s_t)$, we reduced the optimization of Equation 3 to minimizing the L2-norm of the pose errors, resulting in the following loss function:
\begin{equation}
    \mathcal{L}=\sum_{n=1}^N \alpha||t-\hat{t}|| + \beta||r-\hat{r}||,
\end{equation}
where $t$ and $\hat{t}$ are the ground-truth and predicted translation. $r$ and $\hat{r}$ are the ground-truth and predicted rotation. We used Euler angles as the quantitative rotation measure. $\alpha$ and $\beta$ are the translation and rotation error weights, respectively, which were set to 1 and 100 for KITTI and 100 and 20 for EuRoC empirically. We predicted the mean and variance of the stochastic representation $s_t$ and set the minimum variance to be 0.01 to avoid a trivial solution.  We set $\gamma$ in Equation 1 to balance the bottleneck effect. All models were trained for 300 epochs using mini-batches of 16 clips containing five frames each. We set an initial learning rate to 1e-4, which was reduced to 1e-5 and 5e-6 at epoch 150 and 250 to stabilize the training process.

We trained and evaluated the odometry model in a clip-wise manner. For evaluation, we used a sliding window strategy s.t. the evaluated clips are overlapped, which means a frame-pair can appear at different positions in a clip. A refinement strategy that eliminates the results from the first position and averagely ensembles the rest was designed based on our empirical observations, which will be discussed in Section 5.3. Following \cite{sturm2012a} and \cite{chen2019selective}, the averaged root mean squared errors (RMSEs) were used for evaluating both translation and rotation performance. 

\textbf{Remark I:}
In odometry learning, we usually use Euler angles or quaternions for rotation representation rather than SO(3) as implied in SE(3) due to the redundant parameters of the rotation matrix and the orthogonal constraint. We adopt Euler angles in our experiments and assume a Gaussian distribution in this vector space. Though 3D von Mises-Fisher distribution and 4D-Bingham distribution can be arguably more appropriate to model Euler angles and quaternions, respectively, it is non-trivial to evaluate and use them for training in practice.

\textbf{Remark II:}
In terms of the choice of hyperparameters like $\alpha$, $\beta$, and $\gamma$, we basically followed the initial setup of prior works such as \citet{wang2017deepvo,chen2019selective,hafner2020dream} and performed a non-intensive and small-range grid searching. More elegant methods such as relying on the covariance estimates~\citep{peretroukhin2017dpc} can be considered in future studies and applications to new datasets.

\begin{table}
	\caption{Test results on KITTI and EuRoC. We report the averaged RMSEs for translation and rotation, respectively. $\dagger$: Results of MSCKF on KITTI and OKVIS on EuRoC are from \cite{chen2019selective}.} 
	\centering
	\begin{tabular}{c c c c c }
		\specialrule{0.1em}{8pt}{3pt}
		\multirow{2}*{Model} & \multicolumn{2}{c}{KITTI} & \multicolumn{2}{c}{EuRoC} \\
		\cline{2-5}
		\specialrule{0em}{1pt}{1pt}
		~ & $t(m)$ &  $r(^o)$ & $t(m)$ &  $r(^o)$ \\ 
		\specialrule{0.1em}{1pt}{3pt}
		DeepVO & 0.0658 & 0.0942 & 0.0323 & 0.2114 \\
    		\textbf{InfoVO} & \textbf{0.0607} & \textbf{0.0869} & \textbf{0.0310} & \textbf{0.2061} \\
		\specialrule{0.1em}{1pt}{3pt}
		MSCKF/OKVIS$^\dagger$ & 0.116 & 0.044 & 0.0283 & \textbf{0.0402} \\
		\midrule
		VINet & 0.0629 & 0.0453 & 0.0281 & 0.0729 \\
		SoftFusion & 0.0629 & 0.0517 & 0.0281 & 0.0672 \\
		HardFusion & 0.0618 & 0.0447 & 0.0285 & 0.0740 \\
		\midrule
		\textbf{InfoVIO} & 0.0580 & \textbf{0.0416} & 0.0276 & 0.0744 \\
		\textbf{SoftInfoVIO} & 0.0618 & 0.0438 &\textbf{0.0272} & 0.0743 \\
		\textbf{HardInfoVIO} & \textbf{0.0559} & 0.0454 & 0.0291 & 0.0763 \\
		\specialrule{0.1em}{1pt}{3pt}
	\end{tabular}
	\label{tb:main-result}
\end{table}

\subsection{Main Results}
We implemented our visual-inertial framework using three fusion strategies proposed in \cite{chen2019selective}, namely InfoVIO, SoftInfoVIO, and HardInfoVIO. We also included two traditional visual-inertial odometry methods for comparison, i.e., OKVIS~\citep{leutenegger2015keyframe} for EuRoC and MSCKF~\citep{mourikis2007multi,hu2014a} for KITTI. OKVIS is not used for KITTI due to the lack of accurate time synchronization between images and IMU data. Following \cite{sturm2012a} and \cite{chen2019selective}, we report the averaged root mean squared errors (RMSEs) of translation and rotation. The results are given in Table~\ref{tb:main-result}. Our results support the effectiveness of IB w.r.t. the generalizability to test data. Specifically, our basic models (InfoVO/InfoVIO) outperformed all baselines w.r.t. both metrics on KITTI and the translation error on EuRoC. Visual odometry models performed well for translation prediction while incorporating IMU significantly improved the rotation results. Since the MAV trajectories are challenging w.r.t. rotation, the traditional method (OKVIS) still outperformed the other methods, although our result was competitive with the other learning-based baselines. Our re-implementation achieved a better result on KITTI compared with \cite{chen2019selective} but the performance on EuRoC degraded. EuRoC by its nature is much more challenging than KITTI. The major difficulties include (1) the diverse scenarios including an industrial machine hall and an office room, compared with the similar-looking street views in KITTI, (2) the varying difficulty levels of different sequences by manually adjusted obstacles, and (3) the grey-scale images while the FlowNet encoder was pretrained using RGB images, which indicates a domain gap from RGB to grey images and thus degrades the results accordingly. Therefore, reducing the performance gap on EuRoC may require more carefully designed training strategies. Comparisons between the two datasets are summarized in the Supplementary Material. 

\red{
\subsubsection{Visualization of KITTI trajectories}
We further provide per sequence result and trajectory visualization for DeepVO, InfoVO, VINet and InfoVIO to illustrate the benefit of optimizing the IB objective. 

\begin{table}[ht!]
	\caption{\red{Per sequence results on KITTI. We report the averaged translation RMSE drift $t_{rel}$ ($\%$)  on length of 100m-800m and the averaged rotation RMSE drift $r_{rel}$ ($^o/100m$) on length of 100m-800m.}}
	\centering
	\begin{tabular}{c c c c c c c}
		\specialrule{0.1em}{8pt}{3pt}
		\multirow{2}*{Model} & \multicolumn{2}{c}{05} & \multicolumn{2}{c}{07} & \multicolumn{2}{c}{10} \\
		\cline{2-7}
		\specialrule{0em}{1pt}{1pt}
		~ & $t_{rel}$ &  $r_{rel}$ & $t_{rel}$ &  $r_{rel}$ & $t_{rel}$ &  $r_{rel}$ \\ 
		\specialrule{0.1em}{1pt}{3pt}
		DeepVO & 6.25 & 2.29 & 5.66 & 3.60 & 7.12 & \textbf{1.91} \\
		InfoVO & \textbf{4.30} & \textbf{1.54} & \textbf{4.52} & \textbf{3.34} & \textbf{6.25} & 2.16 \\
		\midrule
		VINet  & 3.52 & 1.08 & 5.39 & 3.43 & 8.58 & 2.89 \\
		InfoVIO & \textbf{3.33} & \textbf{0.91} & \textbf{4.69} & \textbf{3.00} & \textbf{7.43} & \textbf{2.44} \\
		\specialrule{0.1em}{1pt}{3pt}
	\end{tabular}
	\label{tb:kitti_per}
\end{table}

\begin{figure}
    \centering
    \includegraphics[width=.5\textwidth]{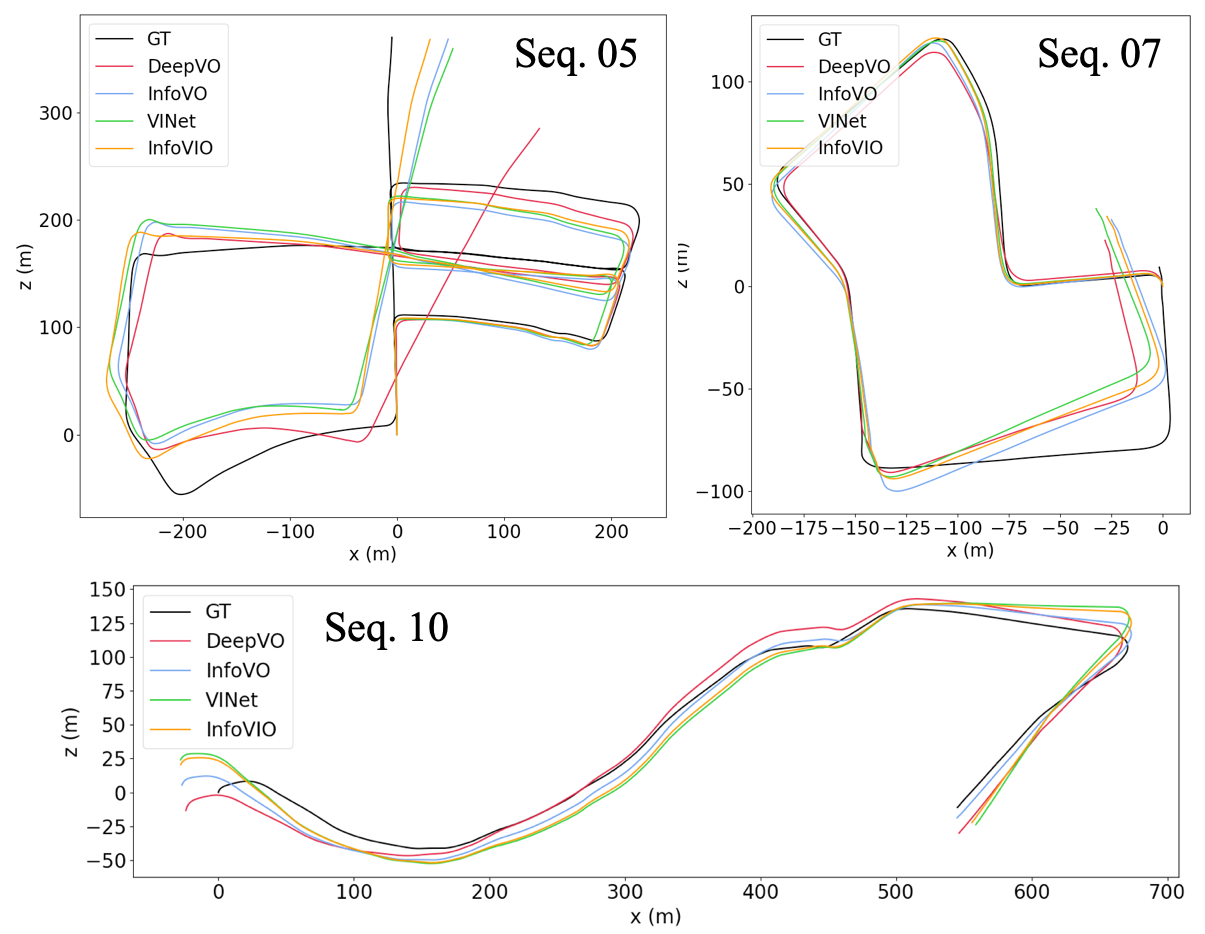}
    \caption{\red{Predicted Trajectories of DeepVO, InfoVO, VINet, and InfoVIO on KITTI sequences 05, 07 and 10.}}
    \label{fig:kitti_traj}
\end{figure}

Results of the test sequences 05, 07, and 10 are presented in Table~\ref{tb:kitti_per} and Fig.~\ref{fig:kitti_traj}. Though long-term accumulated drifts are observed for all end-to-end learning-based odometry methods, InfoVO and InfoVIO that optimize the IB objective still perform better than DeepVO and VINet, especially on sequence 05, which is longer and more challenging due to the increased number of turns. 

}

\red{
\subsection{Generalization to challenging scenarios}
In addition to the results reported on the test splits of KITTI and EuRoC, we further examine the performance of InfoVO on vKITTI2~\citep{cabon2020virtual}, a simulated autonomous driving dataset that contains various scenarios. We illustrate the benefit of the IB objective by training DeepVO and InfoVO on the clean sequences in vKITTI2 and comparing their performance on the more challenging counterparts that have different weather conditions (rain and fog) and lighting conditions (morning, sunset, and overcast). We used Scene 01, 02, and 06 as the training set and left Scene 18 and 20 as the test set. Of note is that only the clean sequences in the training set are used during training. 

\begin{table}[h!]
    \caption{\red{Results on challenging sequences on vKITTI2. W and L denotes sequences that contain different weather conditions (rain and fog) and lighting conditions (morning, sunset, and overcast), respectively.}}
    \centering
	\begin{tabular}{c c c c}
		\specialrule{0.1em}{3pt}{3pt}
		Model & Conditions  & $t(m)$ &  $r(^o)$  \\ 
		\specialrule{0.1em}{1pt}{3pt}
	    DeepVO & W & 1.5214 & 0.1676 \\
		InfoVO & W & 1.5011 & 0.1368 \\
		\specialrule{0.1em}{1pt}{3pt}
		DeepVO & L & 1.4642 & 0.1524 \\
		InfoVO & L & 1.3614 & 0.1239 \\
		\specialrule{0.1em}{1pt}{3pt}
	\end{tabular}
	\label{tb:vkitti2-challenging}
\end{table}

Results under different weather and lighting conditions are presented in Table~\ref{tb:vkitti2-challenging}. It is shown that InfoVO achieves better generalization results in the challenging scenarios than DeepVO w.r.t. both translation and rotation predictions. Besides, our results suggest extreme weather conditions present more challenging than different lighting conditions due to the noises and texture losses in the frames, which remains an interesting research direction towards a more robust odometry system in those challenging scenarios.

}

\red{
\subsection{Compactness of the latent space}
A key hypothesis underlying the motivation to develop our framework is that methods without specific consideration on the compactness of the latent space will implicitly encode pose-irrelevant information into the learnt features, which can be eliminated by the information bottleneck objective. We empirically demonstrated this phenomenon by comparing the reconstruction accuracies using the features learnt by DeepVO and InfoVO. 

Since the optical flow features from the pretrained FlowNet2S~\citep{ilg2017flownet} are used as the network inputs for both DeepVO and InfoVO, we proposed to empirically measure the amount of pose-irrelevant information by the ability to reconstruct those optical flow features from the latent space of DeepVO and InfoVO, respectively. Specifically, we used three MLP layers as the reconstruction decoder, which takes the latent features from the DeepVO and InfoVO models trained on the KITTI dataset as input. We varied the hidden size $d$ of the decoder to examine the performance under different reconstruction capacities. We adopted the same training/test split as in our main experiment and trained the decoder for 300 epochs.

\begin{table}[h!]
    \caption{\red{Results of the reconstruction of optical flow features on KITTI.}}
    \centering
	\begin{tabular}{c c c }
		\specialrule{0.1em}{3pt}{3pt}
		Model & $d$ &  $\bar{l}$  \\ 
		\specialrule{0.1em}{1pt}{3pt}
		DeepVO & 1024 & 0.0387  \\
		DeepVO & 512  & 0.0391  \\
		DeepVO & 256  & 0.0396 \\
		DeepVO & 128  & 0.0401 \\
		\specialrule{0.1em}{1pt}{3pt}
		InfoVO & 1024 & 0.0444  \\
		InfoVO & 512  & 0.0456  \\
		InfoVO & 256  & 0.0508 \\
		InfoVO & 128  & 0.0530 \\
		\specialrule{0.1em}{1pt}{3pt}
		Noise $\sim N(0,1)$ & 1024 & 0.0541  \\
		Noise $\sim N(0,1)$ & 512  & 0.0541  \\
		Noise $\sim N(0,1)$ & 256  & 0.0541 \\
		Noise $\sim N(0,1)$ & 128  & 0.0541 \\
		\specialrule{0.1em}{1pt}{3pt}
	\end{tabular}
	\label{tb:pose-ir}
\end{table}

The results of the averaged MSE loss $\bar{l}$ for optical flow feature reconstruction using different hidden sizes are presented in Table~\ref{tb:pose-ir}. We also reported the results by taking white Gaussian noise as input. It can be seen that DeepVO performs much better than InfoVO in this task, indicating that more pose-irrelevant information that corresponds to optical flows have been encoded by DeepVO implicitly. InfoVO eliminates such information while achieving better pose predictions, as shown in our main results. It is worth noting that the reconstruction performance of InfoVO is close to that of random noise using the hidden size 128, which means although a certain degree of pose-irrelevant information may still exist in the feature space of InfoVO, the remaining amount is small, and it requires a relatively powerful decoder to extract this information.

}

\red{
\subsection{Growing rate of the latent dimension}
As suggested in Corollary~\ref{corollary:2}, the growing rate of the latent dimension $d$ should not exceed that of $n/log(n)$ to avoid overfitting and achieve a tighter generalization bound. To illustrate this effect, we use different sample size ratios for sequence 01 to train InfoVO, and test the trained models on sequences 09 and 10 that have quite different motion patterns (slower vehicle speed) with sequence 01. We first choose the sample size ratio $r_0=1/4$ as the starting point, and empirically determine its corresponding latent dimension $d_0=384$ that leads to neither underfitting nor overfitting. Then we study the performance of InfoVO models using different latent dimensions under the sample size ratios $r_1=1/2$ and $r_2=1.0$, whose growing rates of $n/log(n)$ are $1.780$ and $3.208$, respectively. The results are presented in Fig.~\ref{fig:latent-dim}.

\begin{figure}[ht!]
    \centering
    \includegraphics[width=.45\textwidth]{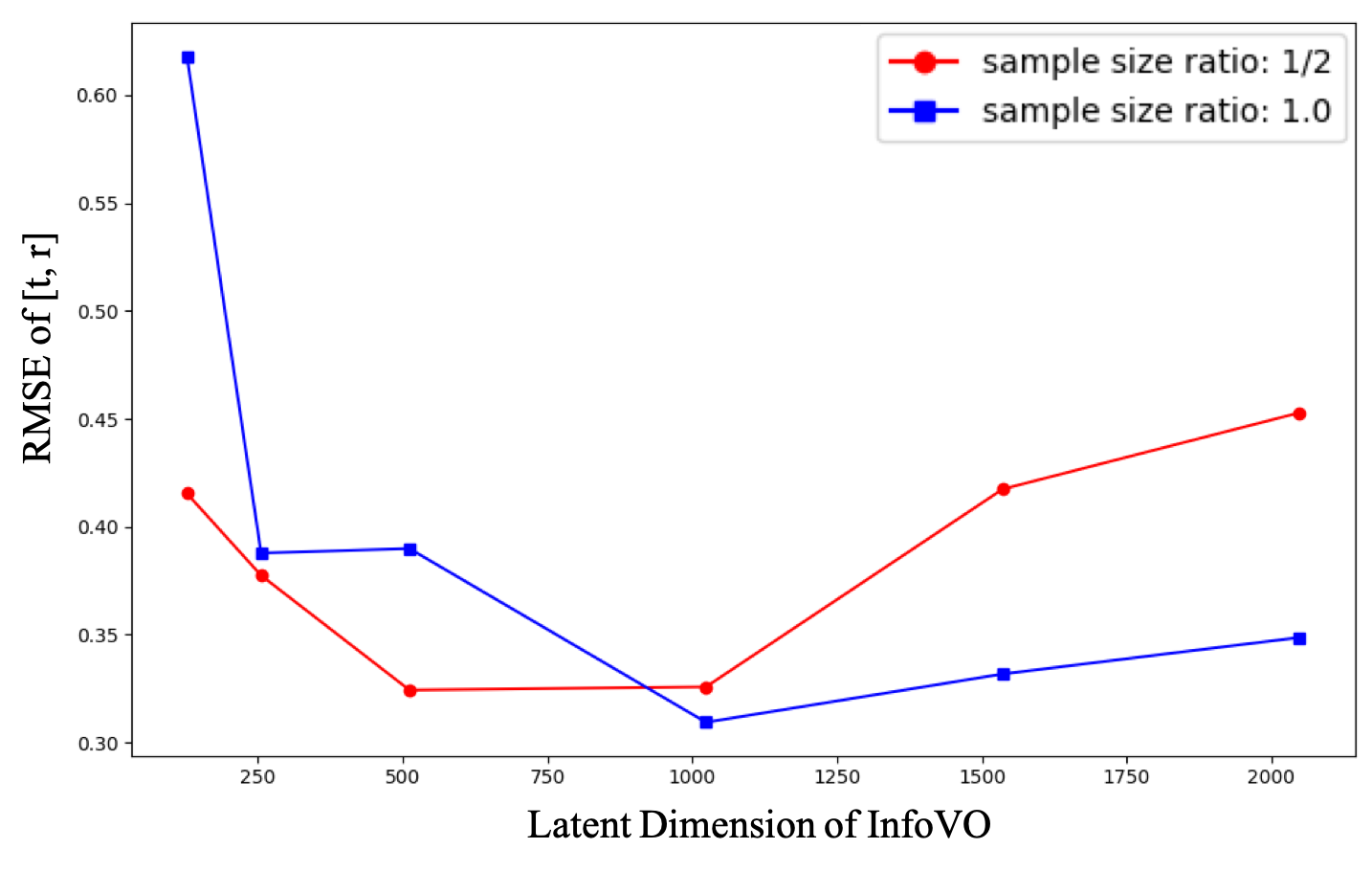}
    \caption{\red{Results of varying latent dimensions (256, 512, 1024, 1536, 2048) under the sample size ratios 1/2 (red) and 1.0 (blue). The RMSE results of the combined 6-DOF translation and rotation vector are reported.}}
    \label{fig:latent-dim}
\end{figure}

We examine the results of latent dimensions 256, 512, 1024, 1536, and 2048. For $r_1=1/2$ and $r_2=1.0$, the latent dimensions that have the same growing rates as $n/log(n)$ are $384 * 1.780\approx684$ and $384 * 3.208\approx1232$, respectively. Accordingly, our results showed that the latent dimensions 512 and 1024 achieved the best test results before overfitting for $r_1=1/2$ and $r_2=1.0$, respectively. A small latent dimension led to an underfitted model while overfitting was observed when the growing rate of the latent dimension exceeds that of $n/log(n)$, which supports Corollary~\ref{corollary:2} empirically.

}

\subsection{Ablation studies}
Extensive ablation studies were conducted to examine the effects of (1) the deterministic component, (2) the IB weight, (3) the sample size and (4) extra sensors. Key observations include: (1) Without the deterministic component, both translation and rotation performance dropped significantly; (2) Determining the IB weight $\gamma$ presents a trade-off between the accuracy of translation and rotation prediction; (3) A larger sample size reduces both the uncertainty and prediction errors; and (4) IMU is more `useful' than cameras for rotation prediction while cameras are more crucial than IMU for translation prediction, according to the discussions on Theorem 2. 

\subsubsection{Effect of the deterministic component}
We conducted stochastic-only ablation experiments to examine the effects of the deterministic components in Equation~\ref{eq:trans_obs} and Equation~\ref{eq:trans_pose} by removing the deterministic nodes in Fig.~\ref{fig:framework_1}(b). We implemented two versions depending on whether the observation- and pose-level latent representations ($s^o$ and $s^p$) were both used as the recurrent network state (StochasticVO/VIO-d), or not (StochasticVO/VIO-s). Results are summarized in Table~\ref{tb:stochastic-only}. Without the deterministic component, both translation and rotation performance dropped significantly, which supports the effectiveness of the proposed deterministic component.

\begin{table}[h!]
    \caption{\iclr{Results of the stochastic-only models on KITTI.}}
    \centering
	\begin{tabular}{c c c }
		\specialrule{0.1em}{3pt}{3pt}
		Model & $t(m)$ &  $r(^o)$  \\ 
		\specialrule{0.1em}{1pt}{3pt}
		StochasticVO-s & 0.0758 & 0.0931  \\
		StochasticVO-d & 0.0783 & 0.0899  \\
		InfoVO (full) & 0.0607 & 0.0869 \\
		\specialrule{0.1em}{1pt}{3pt}
		StochasticVIO-s & 0.0714 & 0.0512  \\
		StochasticVIO-d & 0.0734 & 0.0507  \\
 		InfoVIO (full) & 0.0580 & 0.0416 \\
		\specialrule{0.1em}{1pt}{3pt}
	\end{tabular}
	\label{tb:stochastic-only}
\end{table}

\textbf{Remark:}
For the stochastic-only models, we remove the stochastic state estimators and let the GRU layer in the deterministic transition models directly output the means and standard error residues of the stochastic representation. For state transitions, we then used sampled states as the transitioned state context for the transition model at the next time step. More details of the two implementations are given below. StochasticVO/VIO-d is short for "stochastic VO/VIO with double transition states", which used $(s^o_{t-1},s^p_{t-1})$ as the transition state from the last time step for both observation- and pose-level transitions. StochasticVO/VIO-s is short for "stochastic VO/VIO with single transition states", which used $(s^o_{t-1},s^o_{t-1})$ and $(s^p_{t-1},s^p_{t-1})$  as the transition state from last time step for  observation- and pose-level transitions, respectively.

\subsubsection{Effect of the IB weight}
We examined the effect of the IB weight, i.e. $\gamma$ in Equation~\ref{eq:ib_objective} and Equation~\ref{eq:bottleneck_weight}. \red{As shown in Table~\ref{tb:ib_weight}, Although $\gamma=0.1$ presents a good choice for training on the EuRoC dataset}, we observed that the translation and rotation results did not change consistently with different IB weights \red{on the KITTI dataset.} While the translation accuracy degrades under a larger $\gamma$, the rotation result improves instead. This finding indicates that the determination of the IB weight actually presents a trade-off between the accuracy of translation and rotation predictions and should be taken into account in different scenarios according to the requirements of specific tasks.

\begin{table}[h!]
    \caption{\red{Results of varying IB weights $\gamma$ for InfoVIO.}}
    \centering
	\begin{tabular}{c c c c c}
		\specialrule{0.1em}{2pt}{3pt}
		\multirow{2}*{$\gamma$} & \multicolumn{2}{c}{KITTI} & \multicolumn{2}{c}{EuRoC} \\
		\cline{2-5}
		\specialrule{0em}{1pt}{1pt}
		~ & $t(m)$ &  $r(^o)$ & $t(m)$ &  $r(^o)$ \\ 
		\specialrule{0.1em}{1pt}{3pt}
		$0.0$  & 0.0639 & 0.0482 & 0.0278 & 0.0814 \\
		$0.01$ & 0.0559 & 0.0449 & 0.0277 & 0.0794\\
		$0.05$ & 0.0570 & 0.0424 & 0.0283 & 0.0785 \\
		$0.1$  & 0.0580 & 0.0416 & 0.0276 & 0.0744 \\
		$0.5$  & 0.0612 & 0.0411 & 0.0335 & 0.0765 \\
		$1.0$  & 0.0648 & 0.0375 &  0.0873 & 0.0948 \\
		\specialrule{0.1em}{2pt}{3pt}
	\end{tabular}
	\label{tb:ib_weight}
\end{table}

\subsubsection{Effect of the sample size}
We study the effect of the sample size by using different ratios $r_n$ of training samples for training the model. Recall that we let the minimum variance be 0.01 to avoid a trivial solution, which sets an empirical lower bound of the uncertainty. Table~\ref{tb:uncertainty_sample} shows that a larger sample size reduces both the uncertainty and prediction errors. An interesting observation from our results is that though more training samples still benefit the prediction performance, the averaged variance or the uncertainty measure does not reduce after half of the dataset is added. We suspect that this may be due to the fact that KITTI sequences exhibit quite similar patterns (mostly road driving scenarios). Thus half samples are sufficient for the model to be "familiar" with the dataset and reach the uncertainty margin. While if the training samples are not sufficient enough, e.g. $1/4$ of total samples, the variance increases significantly.

\begin{table}[h!]
    \caption{\iclr{Results of varying sample sizes on KITTI. $r_n$: the ratio of training samples. $\bar{\sigma}^2$: the averaged variance of the latent representation.}}
    \centering
	\begin{tabular}{c c c c}
		\specialrule{0.1em}{2pt}{3pt}
		$r_n$ & $t(m)$ &  $r(^o)$ & $\bar{\sigma}^2$ \\ 
		\specialrule{0.1em}{1pt}{3pt}
		$1/4$ & 0.1977 & 0.1040 & 0.0109 \\
		$1/2$ & 0.0602 & 0.0644 & 0.0101 \\
		$3/4$ & 0.0589 & 0.0544 & 0.0102 \\
		$full$ & 0.0580 & 0.0416 & 0.0102 \\
		\specialrule{0.1em}{2pt}{3pt}
	\end{tabular}
	\label{tb:uncertainty_sample}
\end{table}

\subsubsection{Effect of extra sensors}
Motivated by Theorem~\ref{theorem:2} and our failure-awareness analysis, we study the performance gain of IMU given images and vice versa. The comparison between InfoVO and InfoVIO provides the performance gain of IMU given images. Similarly, to study the performance gain of images given IMU, We trained an IMU-only model, denoted as InfoIO, which is then compared with InfoVIO. The results are summarized in Table~\ref{tb:sensor-gain}, which implies that IMU is more `useful' than cameras for rotation prediction while cameras are more crucial than IMU for translation prediction. Moreover, IMU provides a larger performance gain in EuRoC than KITTI, which is consistent with the fact that the synchronization in EuRoC between IMU and ground-truth poses are more accurate. We also observed that InfoIO performs poorly in KITTI. The large performance gain of images given IMU in KITTI w.r.t. both translation and rotation might also result from the inaccurate alignment of IMU records from the raw KITTI dataset to the image and ground-truth pose sequences.

\begin{table}[h!]
	\caption{Performance gain of IMU given images and images given IMU.} 
	\centering
	\begin{tabular}{c c c c c }
		\specialrule{0.1em}{2pt}{3pt}
		\multirow{2}*{Model} & \multicolumn{2}{c}{KITTI} & \multicolumn{2}{c}{EuRoC} \\
		\cline{2-5}
		\specialrule{0em}{1pt}{1pt}
		~ & $t(m)$ &  $r(^o)$ & $t(m)$ &  $r(^o)$ \\ 
		\specialrule{0.1em}{1pt}{3pt}
		InfoIO & 0.2069 & 0.1164 & 0.0667 & 0.0740 \\
		InfoVO & 0.0607 & 0.0869 & 0.0310 & 0.2061 \\
		InfoVIO & 0.0580 & 0.0416 & 0.0276 & 0.0744 \\
		\specialrule{0.1em}{2pt}{3pt}
	\end{tabular}
	\label{tb:sensor-gain}
\end{table}

\subsection{What Does the Intrinsic Uncertainty Mean?}
We next used the averaged variance of the stochastic latent representation as an intrinsic uncertainty measure and empirically showed how this uncertainty reveals the system properties and data degradation. We found some interesting relationships between the uncertainty and poses, e.g., larger turning angles and smaller forward distances lead to higher uncertainty. Our analysis suggests a practical data collection guideline, i.e., augmenting the uncertain parts of the pose distribution. 

\subsubsection{Uncertainty on KITTI and EuRoC}
\begin{figure*}
    \centering
    \includegraphics[width=.8\textwidth]{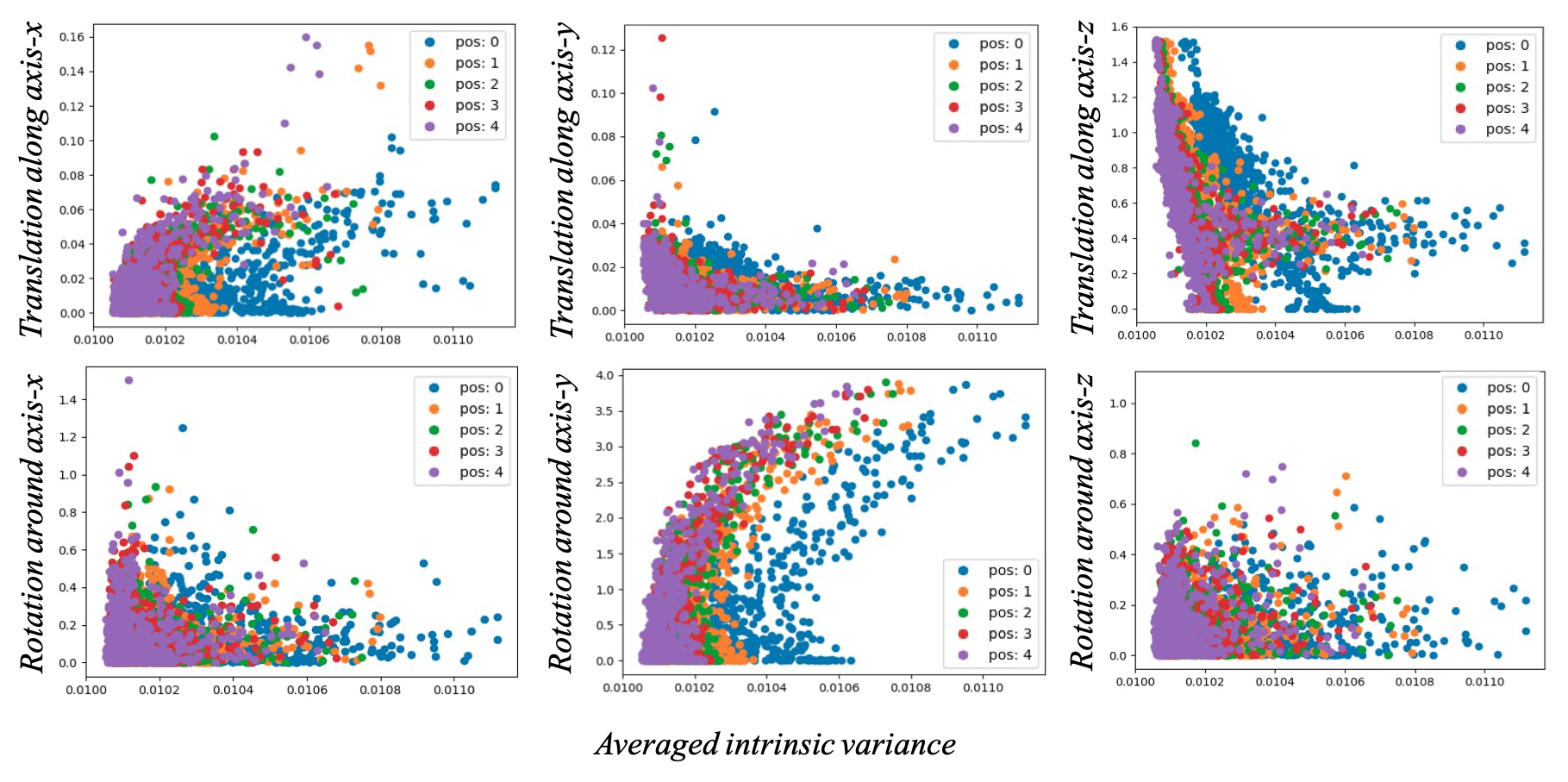}
    \caption{Uncertainty results of InfoVIO on KITTI. The top and bottom rows represent translation and rotation results. The first, second, and third columns represent $x$, $y$, and $z$, respectively. $x,y,z$ are with respect to the coordinate system in KITTI. pos-$i$ means the result is evaluated at the $i$-th position in a clip.}
    \label{fig:uncertainty_kitti}
\end{figure*}

\begin{figure*}
    \centering
    \includegraphics[width=.8\textwidth]{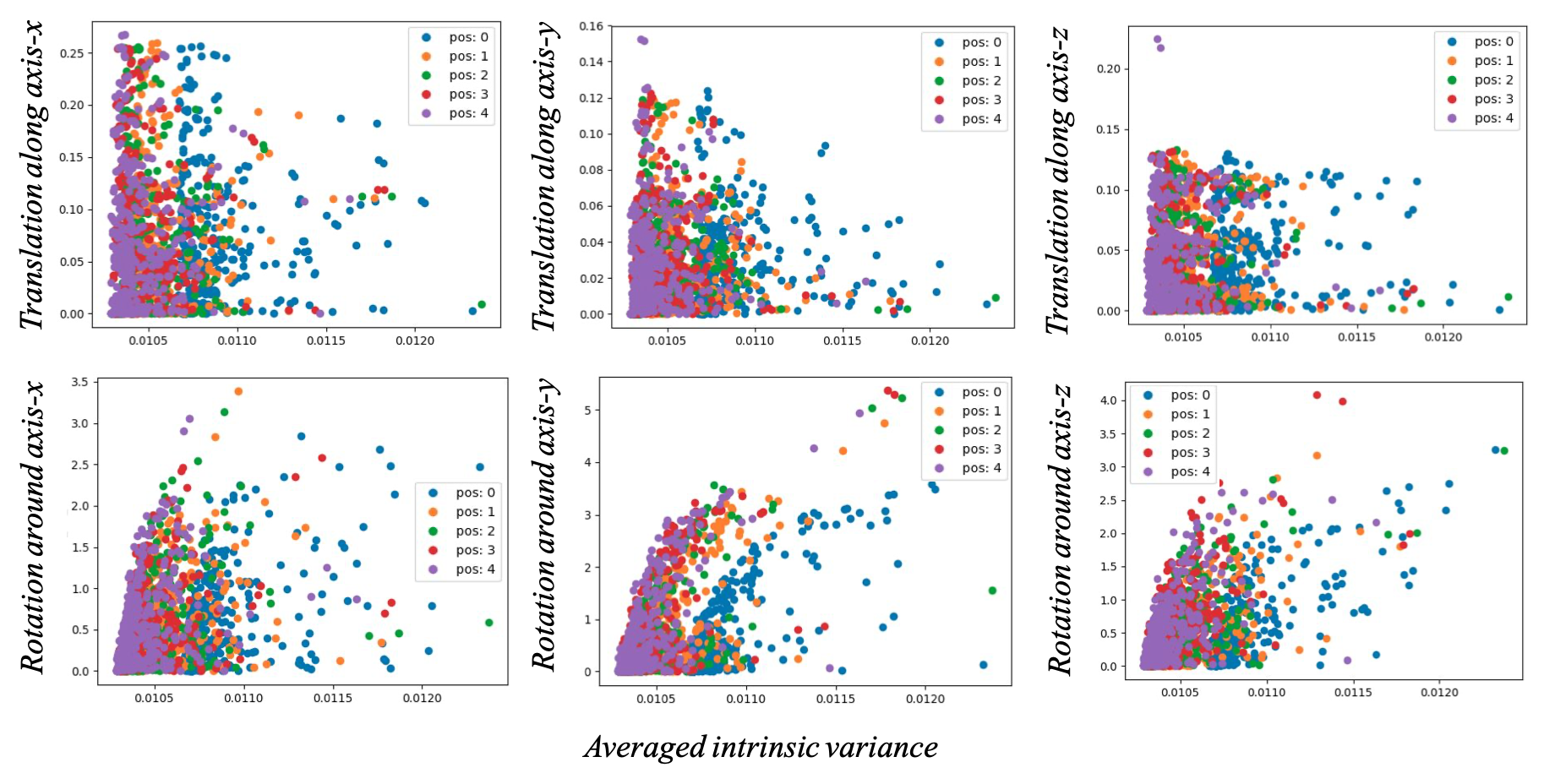}
    \caption{Uncertainty results of InfoVIO on EuRoC. The arrangement and notation are kept the same as Fig.~\ref{fig:uncertainty_kitti}.}
    \label{fig:uncertainty_euroc}
\end{figure*}

We show the uncertainty results of InfoVIO on KITTI and EuRoC in Fig.~\ref{fig:uncertainty_kitti} and Fig.~\ref{fig:uncertainty_euroc}, respectively. Since the translations along $x$ and $y$ axes and the rotations around $x$ and $z$ axes are relatively small in the KITTI dataset, their uncertainties do not exhibit a clear pattern. While for the translation along the forward axis-z and the rotation around the upward axis-y (turning left/right), a clear negative and a clear positive relationship are observed for each motion. The reason for this can be that a large forward parallax provides more distinctive matching features for pose prediction, while a large turning angle instead dramatically reduces the shared visible areas and results in difficulties in achieving accurate predictions. For the EuRoC dataset, we observed a consistent positive relationship for all three rotations, which makes sense in that the MAV rotations are more uniformly distributed along the three axes. The negative relationship in the translation results of EuRoC is more obscure than that of KITTI, partly due to the relative difficulties in accurately predicting MAV translations since EuRoC has a much smaller translation scale than KITTI.

\begin{table}
    \caption{Results on KITTI by evaluating at different positions in a clip.}
    \centering
	\begin{tabular}{c c c c c c c }
		\specialrule{0.1em}{2pt}{3pt}
		$t(m)$ & $pos$-0 &  $pos$-1 &  $pos$-2 &  $pos$-3 &  $pos$-4 \\ 
		\midrule
		DeepVO & 0.0734 & 0.0681 & 0.0661 & \textbf{0.0658} & 0.0659  \\
		InfoVO & 0.0689 & 0.0631 & 0.0618 & 0.0608 & \textbf{0.0604}  \\
		VINet & 0.0683 & 0.0645 & 0.0645 & 0.0632 & \textbf{0.0615}  \\
		InfoVIO & 0.0671 & 0.0602 & 0.0586 & 0.0580 & \textbf{0.0572}  \\
		\specialrule{0.1em}{3pt}{3pt}
		$r(^o)$  & $pos$-0 &  $pos$-1 &  $pos$-2 &  $pos$-3 &  $pos$-4  \\ 
		\midrule
		DeepVO & 0.0970 & 0.0949 & \textbf{0.0939} & 0.0940 & 0.0951  \\
		InfoVO & 0.0904 & 0.0881 & 0.0871 & \textbf{0.0869} & 0.0872  \\
		VINet & 0.0463 & 0.0455 & \textbf{0.0454} & \textbf{0.0454} & 0.0456 \\
		InfoVIO & 0.0427 & \textbf{0.0417} & 0.0420 & 0.0420 & 0.0421 \\
		\specialrule{0.1em}{2pt}{3pt}
	\end{tabular}
    \label{tb:ablation_pos}
\end{table}


\textbf{Remark:}
There is also a line of work that attempts to combine learning based methods with geometric pipelines~\citep{peretroukhin2017dpc,yang2020d3vo}, where uncertainty plays an important role by serving as a quality measure to properly weigh the learned results. The recent successful work by \citet{yang2020d3vo} used learned aleatoric uncertainty to integrate learned results into the DVO pipeline and achieves SOTA performance in monocular odometry. Our work makes contribution in that we do not explicitly learn the variance of final prediction, but use the variance of the intrinsic latent state instead as the uncertainty measure, which we empirically show that can capture the epistemic uncertainty as well and holds the potential to provide better fusion guidance. It remains an interesting future research direction to see whether our uncertainty measure can really benefit this hybrid pipeline that combines the merits of both learning and geometric methods.

\begin{figure*}[h!]
    \centering
    \includegraphics[width=0.8\textwidth]{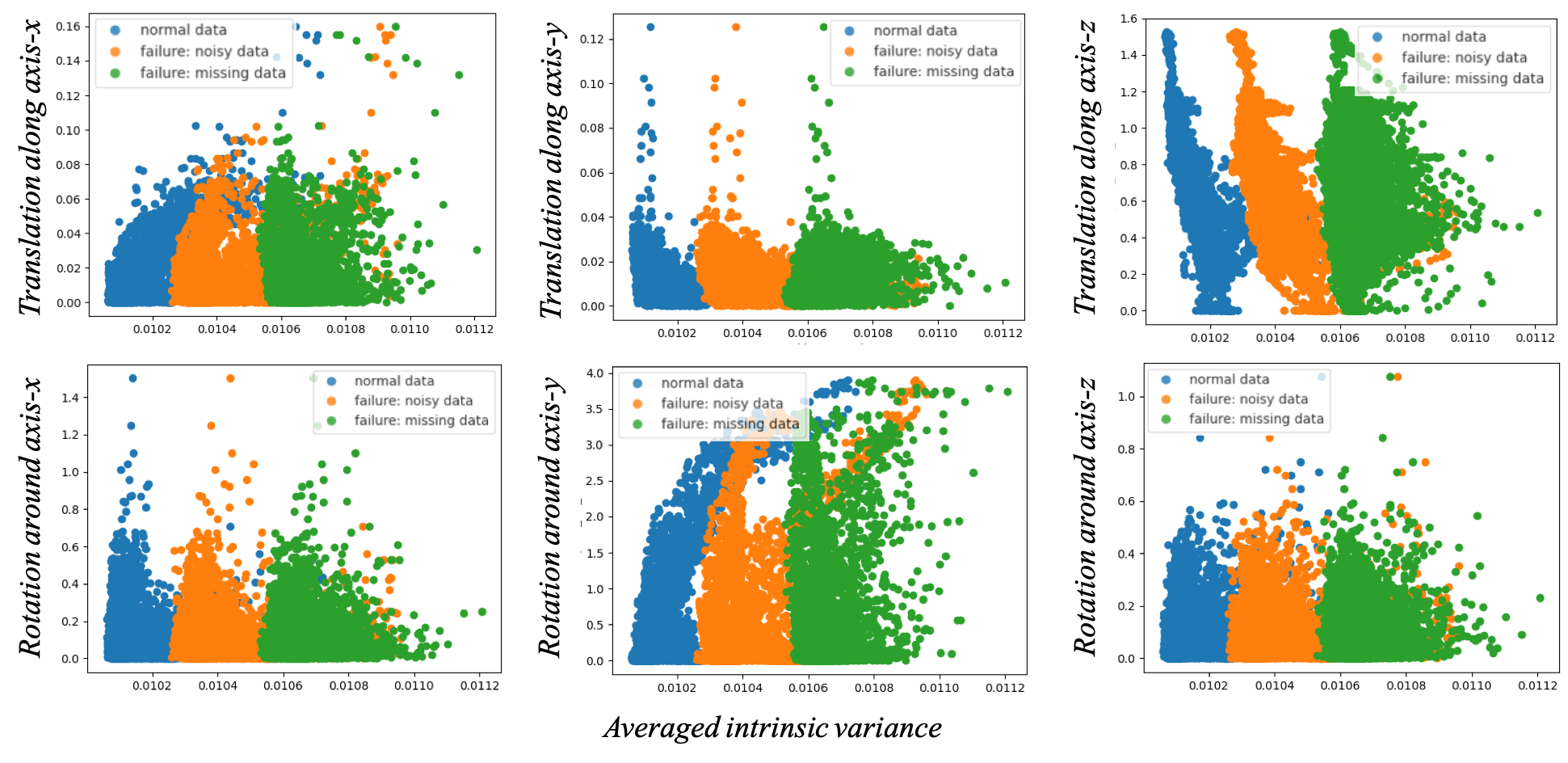}
    \caption{Uncertainty results of InfoVIO on both noisy and missing data of the KITTI dataset. The arrangement and notation are kept the same as Fig.~\ref{fig:uncertainty_kitti}. Blue, orange, and green circles denote results from normal data, noisy data, and missing data, respectively. Both images and IMU records were degraded.}
    \label{fig:uncertainty_mixed}
\end{figure*}

\begin{figure*}[h!]
    \centering
    \includegraphics[width=0.8\textwidth]{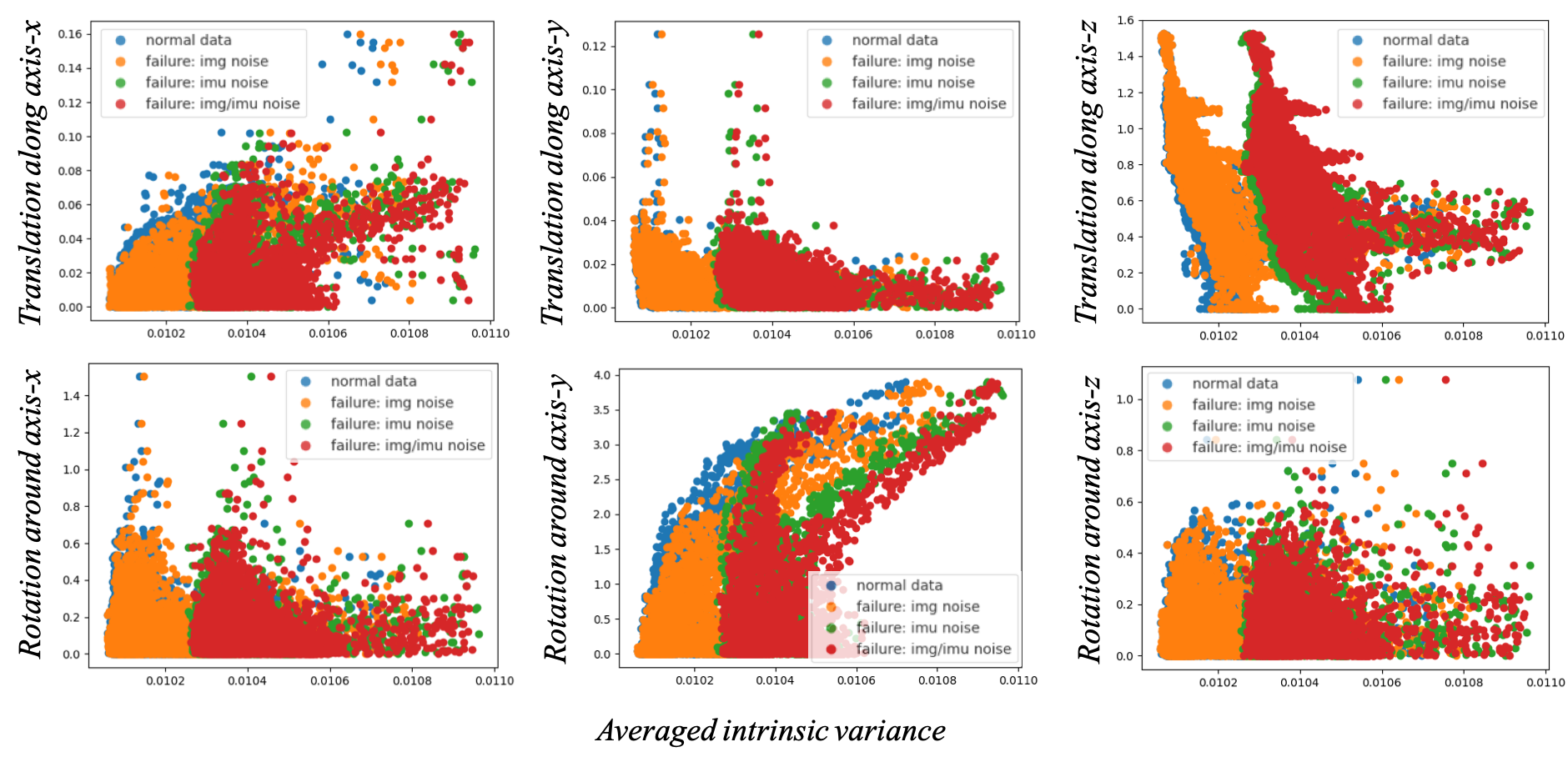}
    \caption{Uncertainty results of InfoVIO on noisy data of the KITTI dataset. The arrangement and notation are kept the same as Fig.~\ref{fig:uncertainty_kitti}. Blue, orange, green, and red circles denote results from normal data and degraded data with images, IMU, and both images and IMU being noisy, respectively.}
    \label{fig:uncertainty_noisy}
\end{figure*}

\begin{figure*}[ht!]
    \centering
    \includegraphics[width=0.8\textwidth]{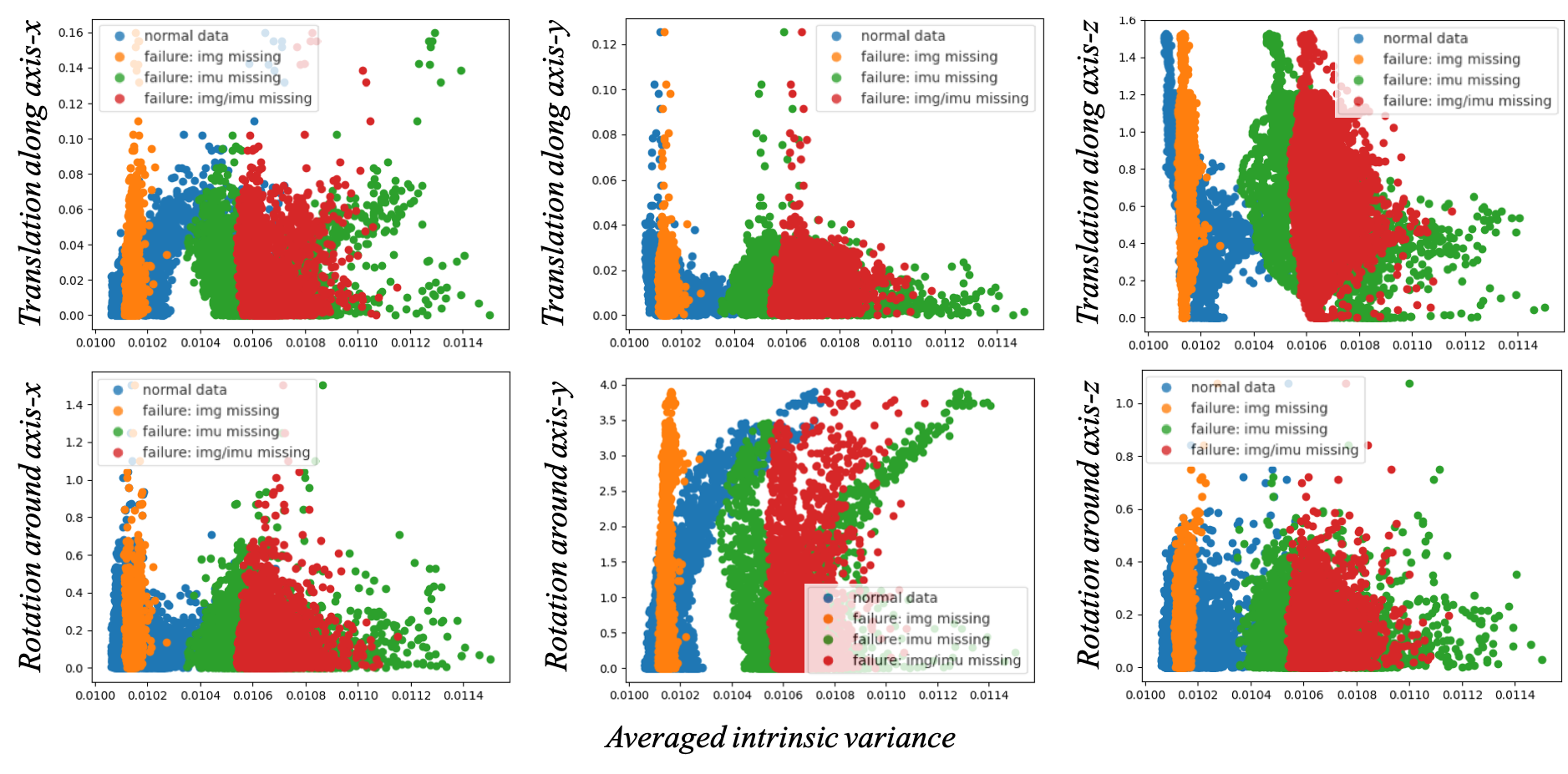}
    \caption{Uncertainty results of InfoVIO on missing data of the KITTI dataset. The arrangement and notation are kept the same as Fig.~\ref{fig:uncertainty_kitti}. Blue, orange, green, and red circles denote results from normal data and degraded data with images, IMU, and both images and IMU missing, respectively.}
    \label{fig:uncertainty_missing}
\end{figure*}

\subsubsection{Uncertainty w.r.t. the evaluated position in a clip} We trained and evaluated the odometry model in a clip-wise manner. Surprisingly, the evaluated position for a frame-pair in consecutive clips also affected the intrinsic uncertainty, as shown in Fig.~\ref{fig:uncertainty_kitti} and Fig.~\ref{fig:uncertainty_euroc}. This makes sense in that when evaluated at a latter position of a clip, the prediction model can leverage more information accumulated from former observations, thus leading to more confident predictions. In Table~\ref{tb:ablation_pos}, we show that, in general, a larger uncertainty results in a higher prediction error. The result also holds for the deterministic DeepVO and VINet baselines, implying that this is a structural system problem in the clip-wise recurrent models. Based on this observation, we propose a simple refinement strategy that eliminates results from the most uncertain position ($pos$-0) and averagely ensembles the results from the rest positions. We report the refined evaluation results for all models in our main results and ablation studies. 

\subsubsection{Failure-awareness}
We show that our intrinsic uncertainty measure is failure-aware, which is crucial for a robust odometry system. We considered two failure cases, namely, degradations with noisy data and missing data. We add Gaussian noise with mean 0 and standard error $0.1$ to the observations in the test dataset to create noisy data. To generate missing data, we replace the observations with the Gaussian noise. 

\begin{table}[h!]
    \caption{\red{Results of the proposed intrinsic uncertainties under different data degradation settings on KITTI and EuRoC. $\bar{\sigma}^2$: the averaged variance of the latent representation. $\checkmark$, $\mathcal{N}$, and $\mathcal{M}$ denote clean, noisy, and missing data, respectively.}}
    \centering
	\begin{tabular}{c c c c c}
		\specialrule{0.1em}{2pt}{3pt}
		 & Image & IMU & $\bar{\sigma}^2$ (KITTI) & $\bar{\sigma}^2$ (EuRoC)\\ 
		\specialrule{0.1em}{1pt}{3pt}
		 Clean & \checkmark & \checkmark & 0.0101 & 0.0103 \\
		 \midrule
		 Noisy & $\mathcal{N}$ & \checkmark & 0.0102 & 0.0103 \\
		 Noisy & \checkmark & $\mathcal{N}$ & 0.0104 & 0.0119 \\
		 Noisy & $\mathcal{N}$ & $\mathcal{N}$ & 0.0104 & 0.0119 \\
		 \midrule 
		 Missing  & $\mathcal{M}$ & \checkmark & 0.0101 & 0.0103 \\
		 Missing  & \checkmark & $\mathcal{M}$ & 0.0106 & 0.0119 \\
		 Missing  & $\mathcal{M}$  & $\mathcal{M}$  & 0.0107 & 0.0119 \\
		\specialrule{0.1em}{2pt}{3pt}
	\end{tabular}
	\label{tb:var_degrade}
\end{table}

\red{In Fig.~\ref{fig:uncertainty_mixed}, we report the visualization results of uncertainties versus different translations and rotations on KITTI by applying data corruption to both images and IMU. The results of single sensor corruption under the noisy and missing data settings are also provided in Fig.~\ref{fig:uncertainty_noisy} and Fig.~\ref{fig:uncertainty_missing}, respectively. The visualization results on EuRoC is provided in the Supplementary Material. We summarize the intrinsic variances under different data degradation settings  in Table~\ref{tb:var_degrade}. Our model becomes more uncertain as the data degrades. The uncertainty reaches the highest when the data is missing, as expected. A more interesting observation is that the quality of IMU data dominates the uncertainty for both KITTI and EuRoC, implying that current image encoders are not trained well enough, and a better image encoder is desirable to fully utilize the visual information. Also, data degradation on IMU records leads to higher uncertainty in EuRoC than in KITTI. We suspect the reason is that the synchronization between the ground-truth poses and IMU records are less accurate in KITTI than in EuRoC, leading to noisy IMU data for training. At last, the model trained on EuRoC exhibits the same performance on the noisy and the missing data, which implies that EuRoC dataset may be more prone to noises. These observations support that the proposed intrinsic uncertainty measure provides a practical tool for failure diagnoses, such as noises, sensor malfunctions, and even mis-synchronization between sensors.}

\section{Conclusion and Future Research}

This paper targets odometry learning by proposing an information-theoretic framework that leverages an IB-based objective function to eliminate the pose-irrelevant information. A recurrent deterministic-stochastic transition model is introduced to facilitate the modeling of time dependency of the observation sequences. The proposed framework can be easily extended to different problem settings and provide not only an intrinsic uncertainty measure but also an elegant theoretical analysis tool for evaluating the system performance. We derive generalization error bounds for the IB-based method and a predictability lower bound for the latent representation given extra sensors. They provide theoretical performance guarantees for the proposed framework, \iclr{and more generally, information-bottleneck based methods}. Extensive experiments on KITTI and EuRoC support our discoveries.

The proposed method falls into end-to-end supervised learning methods. Obtaining the required ground-truth pose labels can be challenging for large-scale data collection and training. Two recent research trends provide promising solutions to mitigate this problem, i.e. embodied methods that utilize simulated environments and unsupervised learning methods that leveraged the geometric constraints and trained the model jointly with other auxiliary tasks like depth prediction. The difficulty in bringing embodied methods into current state-of-the-art frameworks is the domain gap between simulation and the real world, where proper domain adaptation techniques are desired. Integrating unsupervised and supervised methods can also be challenging, which requires more dedicated training strategies and model design. It is worth noting that our proposed IB method improves on the representation level and can also be applied in these fields to obtain better latent representations. We foresee further developments by incorporating novel techniques into our IB framework.

\bibliographystyle{plainnat}
\bibliography{ijcv}   

\begin{thebibliography}{50}
\providecommand{\natexlab}[1]{#1}
\providecommand{\url}[1]{\texttt{#1}}
\expandafter\ifx\csname urlstyle\endcsname\relax
  \providecommand{\doi}[1]{doi: #1}\else
  \providecommand{\doi}{doi: \begingroup \urlstyle{rm}\Url}\fi

\bibitem[{Alemi} et~al.(2017){Alemi}, {Fischer}, {Dillon}, and
  {Murphy}]{alemi2017deep}
Alexander~A. {Alemi}, Ian {Fischer}, Joshua~V. {Dillon}, and Kevin {Murphy}.
\newblock Deep variational information bottleneck.
\newblock In \emph{ICLR 2017 : International Conference on Learning
  Representations 2017}, 2017.

\bibitem[{Bian} et~al.(2019){Bian}, {Li}, {Wang}, {Zhan}, {Shen}, {Cheng}, and
  {Reid}]{bian2019unsupervised}
JiaWang {Bian}, Zhichao {Li}, Naiyan {Wang}, Huangying {Zhan}, Chunhua {Shen},
  Ming-Ming {Cheng}, and Ian {Reid}.
\newblock Unsupervised scale-consistent depth and ego-motion learning from
  monocular video.
\newblock In \emph{NeurIPS 2019 : Thirty-third Conference on Neural Information
  Processing Systems}, pages 35--45, 2019.

\bibitem[{Buesing} et~al.(){Buesing}, {Weber}, {Racanière}, {Eslami},
  {Rezende}, {Reichert}, {Viola}, {Besse}, {Gregor}, {Hassabis}, and
  {Wierstra}]{buesing2018learning}
Lars {Buesing}, Theophane {Weber}, Sébastien {Racanière}, S.~M.~Ali {Eslami},
  Danilo~Jimenez {Rezende}, David~P. {Reichert}, Fabio {Viola}, Frederic
  {Besse}, Karol {Gregor}, Demis {Hassabis}, and Daan {Wierstra}.
\newblock Learning and querying fast generative models for reinforcement
  learning.
\newblock \emph{FAMI workshop}.

\bibitem[{Burri} et~al.(2016){Burri}, {Nikolic}, {Gohl}, {Schneider}, {Rehder},
  {Omari}, {Achtelik}, and {Siegwart}]{burri2016the}
Michael {Burri}, Janosch {Nikolic}, Pascal {Gohl}, Thomas {Schneider}, Joern
  {Rehder}, Sammy {Omari}, Markus~W {Achtelik}, and Roland {Siegwart}.
\newblock The euroc micro aerial vehicle datasets.
\newblock \emph{The International Journal of Robotics Research}, 35\penalty0
  (10):\penalty0 1157--1163, 2016.

\bibitem[Cabon et~al.(2020)Cabon, Murray, and Humenberger]{cabon2020virtual}
Yohann Cabon, Naila Murray, and Martin Humenberger.
\newblock Virtual kitti 2.
\newblock \emph{arXiv preprint arXiv:2001.10773}, 2020.

\bibitem[{Chaudhari} et~al.(2017){Chaudhari}, {Choromanska}, {Soatto}, {LeCun},
  {Baldassi}, {Borgs}, {Chayes}, {Sagun}, and {Zecchina}]{chaudhari2017entropy}
Pratik {Chaudhari}, Anna {Choromanska}, Stefano {Soatto}, Yann {LeCun}, Carlo
  {Baldassi}, Christian {Borgs}, Jennifer {Chayes}, Levent {Sagun}, and
  Riccardo {Zecchina}.
\newblock Entropy-sgd: Biasing gradient descent into wide valleys.
\newblock \emph{international conference on learning representations},
  2019\penalty0 (12):\penalty0 124018, 2017.

\bibitem[{Chen} et~al.(2019){Chen}, {Rosa}, {Miao}, {Lu}, {Wu}, {Markham}, and
  {Trigoni}]{chen2019selective}
Changhao {Chen}, Stefano {Rosa}, Yishu {Miao}, Chris~Xiaoxuan {Lu}, Wei {Wu},
  Andrew {Markham}, and Niki {Trigoni}.
\newblock Selective sensor fusion for neural visual-inertial odometry.
\newblock In \emph{2019 IEEE/CVF Conference on Computer Vision and Pattern
  Recognition (CVPR)}, pages 10542--10551, 2019.

\bibitem[{Chen} et~al.(2020){Chen}, {Wang}, {Lu}, {Trigoni}, and
  {Markham}]{chen2020a}
Changhao {Chen}, Bing {Wang}, Chris~Xiaoxuan {Lu}, Niki {Trigoni}, and Andrew
  {Markham}.
\newblock A survey on deep learning for localization and mapping: Towards the
  age of spatial machine intelligence.
\newblock \emph{arXiv preprint arXiv:2006.12567}, 2020.

\bibitem[{Cho} et~al.(2014){Cho}, van {Merrienboer}, {Gulcehre}, {Bahdanau},
  {Bougares}, {Schwenk}, and {Bengio}]{cho2014learning}
Kyunghyun {Cho}, Bart van {Merrienboer}, Caglar {Gulcehre}, Dzmitry {Bahdanau},
  Fethi {Bougares}, Holger {Schwenk}, and Yoshua {Bengio}.
\newblock Learning phrase representations using rnn encoder--decoder for
  statistical machine translation.
\newblock In \emph{Proceedings of the 2014 Conference on Empirical Methods in
  Natural Language Processing (EMNLP)}, pages 1724--1734, 2014.

\bibitem[{Chung} et~al.(2015){Chung}, {Kastner}, {Dinh}, {Goel}, {Courville},
  and {Bengio}]{chung2015a}
Junyoung {Chung}, Kyle {Kastner}, Laurent {Dinh}, Kratarth {Goel}, Aaron
  {Courville}, and Yoshua {Bengio}.
\newblock A recurrent latent variable model for sequential data.
\newblock In \emph{NIPS'15 Proceedings of the 28th International Conference on
  Neural Information Processing Systems - Volume 2}, pages 2980--2988, 2015.

\bibitem[{Clark} et~al.(2017){Clark}, {Wang}, {Wen}, {Markham}, and
  {Trigoni}]{clark2017vinet}
Ronald {Clark}, Sen {Wang}, Hongkai {Wen}, Andrew {Markham}, and Niki
  {Trigoni}.
\newblock Vinet: Visual inertial odometry as a sequence to sequence learning
  problem.
\newblock In \emph{Thirty-First AAAI Conference on Artificial Intelligence},
  pages 3995--4001, 2017.

\bibitem[{Cover} and {Thomas}(1991)]{cover1991elements}
Thomas~M. {Cover} and Joy~A. {Thomas}.
\newblock \emph{Elements of information theory}.
\newblock 1991.

\bibitem[{Dai} et~al.(2018){Dai}, {Zhu}, and {Wipf}]{dai2018compressing}
Bin {Dai}, Chen {Zhu}, and David {Wipf}.
\newblock Compressing neural networks using the variational information
  bottelneck.
\newblock In \emph{ICML 2018: Thirty-fifth International Conference on Machine
  Learning}, pages 1135--1144, 2018.

\bibitem[{Dosovitskiy} et~al.(2015){Dosovitskiy}, {Fischery}, {Ilg}, {Hausser},
  {Hazirbas}, {Golkov}, van~der {Smagt}, {Cremers}, and
  {Brox}]{dosovitskiy2015flownet}
Alexey {Dosovitskiy}, Philipp {Fischery}, Eddy {Ilg}, Philip {Hausser}, Caner
  {Hazirbas}, Vladimir {Golkov}, Patrick van~der {Smagt}, Daniel {Cremers}, and
  Thomas {Brox}.
\newblock Flownet: Learning optical flow with convolutional networks.
\newblock In \emph{2015 IEEE International Conference on Computer Vision
  (ICCV)}, pages 2758--2766, 2015.

\bibitem[{Durrant-Whyte} and {Bailey}(2006)]{durrant-whyte2006simultaneous}
H.~{Durrant-Whyte} and T.~{Bailey}.
\newblock Simultaneous localization and mapping: part i.
\newblock \emph{IEEE Robotics \& Automation Magazine}, 13\penalty0
  (2):\penalty0 99--110, 2006.

\bibitem[{Fuentes-Pacheco} et~al.(2015){Fuentes-Pacheco}, {Ruiz-Ascencio}, and
  {Rendón-Mancha}]{fuentes-pacheco2015visual}
Jorge {Fuentes-Pacheco}, José {Ruiz-Ascencio}, and Juan~Manuel
  {Rendón-Mancha}.
\newblock Visual simultaneous localization and mapping: a survey.
\newblock \emph{Artificial Intelligence Review}, 43\penalty0 (1):\penalty0
  55--81, 2015.

\bibitem[{Gal} and {Ghahramani}(2016)]{gal2016dropout}
Yarin {Gal} and Zoubin {Ghahramani}.
\newblock Dropout as a bayesian approximation: representing model uncertainty
  in deep learning.
\newblock In \emph{ICML'16 Proceedings of the 33rd International Conference on
  International Conference on Machine Learning - Volume 48}, pages 1050--1059,
  2016.

\bibitem[{Geiger} et~al.(2013){Geiger}, {Lenz}, {Stiller}, and
  {Urtasun}]{geiger2013vision}
A~{Geiger}, P~{Lenz}, C~{Stiller}, and R~{Urtasun}.
\newblock Vision meets robotics: The kitti dataset.
\newblock \emph{The International Journal of Robotics Research}, 32\penalty0
  (11):\penalty0 1231--1237, 2013.

\bibitem[{Goyal} et~al.(2019){Goyal}, {Islam}, {Strouse}, {Ahmed},
  {Larochelle}, {Botvinick}, {Levine}, and {Bengio}]{goyal2019infobot}
Anirudh Goyal Alias~Parth {Goyal}, Riashat {Islam}, DJ~{Strouse}, Zafarali
  {Ahmed}, Hugo {Larochelle}, Matthew {Botvinick}, Sergey {Levine}, and Yoshua
  {Bengio}.
\newblock Infobot: Transfer and exploration via the information bottleneck.
\newblock In \emph{ICLR 2019 : 7th International Conference on Learning
  Representations}, 2019.

\bibitem[{Hafner} et~al.(2019){Hafner}, {Lillicrap}, {Fischer}, {Villegas},
  {Ha}, {Lee}, and {Davidson}]{hafner2019learning}
Danijar {Hafner}, Timothy {Lillicrap}, Ian {Fischer}, Ruben {Villegas}, David
  {Ha}, Honglak {Lee}, and James {Davidson}.
\newblock Learning latent dynamics for planning from pixels.
\newblock In \emph{ICML 2019 : Thirty-sixth International Conference on Machine
  Learning}, pages 2555--2565, 2019.

\bibitem[{Hafner} et~al.(2020){Hafner}, {Lillicrap}, {Ba}, and
  {Norouzi}]{hafner2020dream}
Danijar {Hafner}, Timothy {Lillicrap}, Jimmy {Ba}, and Mohammad {Norouzi}.
\newblock Dream to control: Learning behaviors by latent imagination.
\newblock In \emph{ICLR 2020 : Eighth International Conference on Learning
  Representations}, 2020.

\bibitem[{Hu} and {Chen}(2014)]{hu2014a}
Jwu-Sheng {Hu} and Ming-Yuan {Chen}.
\newblock A sliding-window visual-imu odometer based on tri-focal tensor
  geometry.
\newblock In \emph{2014 IEEE International Conference on Robotics and
  Automation (ICRA)}, pages 3963--3968, 2014.

\bibitem[{Ilg} et~al.(2017){Ilg}, {Mayer}, {Saikia}, {Keuper}, {Dosovitskiy},
  and {Brox}]{ilg2017flownet}
Eddy {Ilg}, Nikolaus {Mayer}, Tonmoy {Saikia}, Margret {Keuper}, Alexey
  {Dosovitskiy}, and Thomas {Brox}.
\newblock Flownet 2.0: Evolution of optical flow estimation with deep networks.
\newblock In \emph{2017 IEEE Conference on Computer Vision and Pattern
  Recognition (CVPR)}, pages 1647--1655, 2017.

\bibitem[{Kendall} and {Gal}(2017)]{kendall2017what}
Alex {Kendall} and Yarin {Gal}.
\newblock What uncertainties do we need in bayesian deep learning for computer
  vision.
\newblock In \emph{NIPS'17 Proceedings of the 31st International Conference on
  Neural Information Processing Systems}, pages 5580--5590, 2017.

\bibitem[{Kendall} et~al.(2015){Kendall}, {Grimes}, and
  {Cipolla}]{kendall2015posenet}
Alex {Kendall}, Matthew {Grimes}, and Roberto {Cipolla}.
\newblock Posenet: A convolutional network for real-time 6-dof camera
  relocalization.
\newblock In \emph{2015 IEEE International Conference on Computer Vision
  (ICCV)}, pages 2938--2946, 2015.

\bibitem[{Kingma} and {Welling}(2014)]{kingma2014auto}
Diederik~P {Kingma} and Max {Welling}.
\newblock Auto-encoding variational bayes.
\newblock In \emph{ICLR 2014 : International Conference on Learning
  Representations (ICLR) 2014}, 2014.

\bibitem[{Leutenegger} et~al.(2015){Leutenegger}, {Lynen}, {Bosse}, {Siegwart},
  and {Furgale}]{leutenegger2015keyframe}
Stefan {Leutenegger}, Simon {Lynen}, Michael {Bosse}, Roland {Siegwart}, and
  Paul {Furgale}.
\newblock Keyframe-based visual-inertial odometry using nonlinear optimization.
\newblock \emph{The International Journal of Robotics Research}, 34\penalty0
  (3):\penalty0 314--334, 2015.

\bibitem[{Loquercio} et~al.(2020){Loquercio}, {Segu}, and
  {Scaramuzza}]{loquercio2020a}
Antonio {Loquercio}, Mattia {Segu}, and Davide {Scaramuzza}.
\newblock A general framework for uncertainty estimation in deep learning.
\newblock \emph{In 2020 IEEE International Conference on Robotics and
  Automation (ICRA)}, 5\penalty0 (2):\penalty0 3153--3160, 2020.

\bibitem[{MacKay}(1992)]{mackay1992a}
David J.~C. {MacKay}.
\newblock A practical bayesian framework for backpropagation networks.
\newblock \emph{Neural Computation}, 4\penalty0 (3):\penalty0 448--472, 1992.

\bibitem[Mourikis and Roumeliotis(2007)]{mourikis2007multi}
Anastasios~I Mourikis and Stergios~I Roumeliotis.
\newblock A multi-state constraint kalman filter for vision-aided inertial
  navigation.
\newblock In \emph{Proceedings 2007 IEEE International Conference on Robotics
  and Automation}, pages 3565--3572. IEEE, 2007.

\bibitem[{Peretroukhin} and {Kelly}(2017)]{peretroukhin2017dpc}
Valentin {Peretroukhin} and Jonathan {Kelly}.
\newblock Dpc-net: Deep pose correction for visual localization.
\newblock \emph{international conference on robotics and automation},
  3\penalty0 (3):\penalty0 2424--2431, 2017.

\bibitem[{Poole} et~al.(2019){Poole}, {Ozair}, van~den {Oord}, {Alemi}, and
  {Tucker}]{poole2019on}
Ben {Poole}, Sherjil {Ozair}, Aäron van~den {Oord}, Alexander {Alemi}, and
  George {Tucker}.
\newblock On variational bounds of mutual information.
\newblock In \emph{ICML 2019 : Thirty-sixth International Conference on Machine
  Learning}, pages 5171--5180, 2019.

\bibitem[{Ranjan} et~al.(2019){Ranjan}, {Jampani}, {Balles}, {Kim}, {Sun},
  {Wulff}, and {Black}]{ranjan2019competitive}
Anurag {Ranjan}, Varun {Jampani}, Lukas {Balles}, Kihwan {Kim}, Deqing {Sun},
  Jonas {Wulff}, and Michael~J. {Black}.
\newblock Competitive collaboration: Joint unsupervised learning of depth,
  camera motion, optical flow and motion segmentation.
\newblock In \emph{2019 IEEE/CVF Conference on Computer Vision and Pattern
  Recognition (CVPR)}, pages 12240--12249, 2019.

\bibitem[{Shwartz-Ziv} and {Tishby}(2017)]{shwartz-ziv2017opening}
Ravid {Shwartz-Ziv} and Naftali {Tishby}.
\newblock Opening the black box of deep neural networks via information.
\newblock \emph{arXiv preprint arXiv:1703.00810}, 2017.

\bibitem[{Steiner} et~al.(2019){Steiner}, {DeVito}, {Chintala}, {Gross},
  {Paszke}, {Massa}, {Lerer}, {Chanan}, {Lin}, {Yang}, {Desmaison}, {Tejani},
  {Kopf}, {Bradbury}, {Antiga}, {Raison}, {Gimelshein}, {Chilamkurthy},
  {Killeen}, {Fang}, and {Bai}]{steiner2019pytorch}
Benoit {Steiner}, Zachary {DeVito}, Soumith {Chintala}, Sam {Gross}, Adam
  {Paszke}, Francisco {Massa}, Adam {Lerer}, Gregory {Chanan}, Zeming {Lin},
  Edward {Yang}, Alban {Desmaison}, Alykhan {Tejani}, Andreas {Kopf}, James
  {Bradbury}, Luca {Antiga}, Martin {Raison}, Natalia {Gimelshein}, Sasank
  {Chilamkurthy}, Trevor {Killeen}, Lu~{Fang}, and Junjie {Bai}.
\newblock Pytorch: An imperative style, high-performance deep learning library.
\newblock In \emph{NeurIPS 2019 : Thirty-third Conference on Neural Information
  Processing Systems}, pages 8026--8037, 2019.

\bibitem[{Sturm} et~al.(2012){Sturm}, {Engelhard}, {Endres}, {Burgard}, and
  {Cremers}]{sturm2012a}
Jrgen {Sturm}, Nikolas {Engelhard}, Felix {Endres}, Wolfram {Burgard}, and
  Daniel {Cremers}.
\newblock A benchmark for the evaluation of rgb-d slam systems.
\newblock In \emph{2012 IEEE/RSJ International Conference on Intelligent Robots
  and Systems}, pages 573--580, 2012.

\bibitem[{Taketomi} et~al.(2017){Taketomi}, {Uchiyama}, and
  {Ikeda}]{taketomi2017visual}
Takafumi {Taketomi}, Hideaki {Uchiyama}, and Sei {Ikeda}.
\newblock Visual slam algorithms: a survey from 2010 to 2016.
\newblock \emph{IPSJ Transactions on Computer Vision and Applications},
  9\penalty0 (1):\penalty0 16, 2017.

\bibitem[{Tishby} and {Zaslavsky}(2015)]{tishby2015deep}
Naftali {Tishby} and Noga {Zaslavsky}.
\newblock Deep learning and the information bottleneck principle.
\newblock In \emph{2015 IEEE Information Theory Workshop (ITW)}, pages 1--5,
  2015.

\bibitem[{Tishby} et~al.(2000){Tishby}, {Pereira}, and {Bialek}]{tishby2000the}
Naftali {Tishby}, Fernando C.~N. {Pereira}, and William {Bialek}.
\newblock The information bottleneck method.
\newblock \emph{Proc. 37th Annual Allerton Conference on Communications,
  Control and Computing, 1999}, pages 368--377, 2000.

\bibitem[{Vera} et~al.(2018){Vera}, {Piantanida}, and {Vega}]{vera2018the}
Matias {Vera}, Pablo {Piantanida}, and Leonardo~Rey {Vega}.
\newblock The role of the information bottleneck in representation learning.
\newblock In \emph{2018 IEEE International Symposium on Information Theory
  (ISIT)}, pages 1580--1584, 2018.

\bibitem[{Wang} et~al.(2017){Wang}, {Clark}, {Wen}, and
  {Trigoni}]{wang2017deepvo}
Sen {Wang}, Ronald {Clark}, Hongkai {Wen}, and Niki {Trigoni}.
\newblock Deepvo: Towards end-to-end visual odometry with deep recurrent
  convolutional neural networks.
\newblock In \emph{2017 IEEE International Conference on Robotics and
  Automation (ICRA)}, pages 2043--2050, 2017.

\bibitem[{Wang} et~al.(2018){Wang}, {Clark}, {Wen}, and {Trigoni}]{wang2018end}
Sen {Wang}, Ronald {Clark}, Hongkai {Wen}, and Niki {Trigoni}.
\newblock End-to-end, sequence-to-sequence probabilistic visual odometry
  through deep neural networks:.
\newblock \emph{The International Journal of Robotics Research}, 37:\penalty0
  513--542, 2018.

\bibitem[{Xu} and {Raginsky}(2017)]{xu2017information}
Aolin {Xu} and Maxim {Raginsky}.
\newblock Information-theoretic analysis of generalization capability of
  learning algorithms.
\newblock In \emph{31st Annual Conference on Neural Information Processing
  Systems, NIPS 2017}, pages 2524--2533, 2017.

\bibitem[{Xue} et~al.(2019){Xue}, {Wang}, {Li}, {Wang}, {Wang}, and
  {Zha}]{xue2019beyond}
Fei {Xue}, Xin {Wang}, Shunkai {Li}, Qiuyuan {Wang}, Junqiu {Wang}, and Hongbin
  {Zha}.
\newblock Beyond tracking: Selecting memory and refining poses for deep visual
  odometry.
\newblock In \emph{2019 IEEE/CVF Conference on Computer Vision and Pattern
  Recognition (CVPR)}, pages 8575--8583, 2019.

\bibitem[Yang et~al.(2020)Yang, Stumberg, Wang, and Cremers]{yang2020d3vo}
Nan Yang, Lukas~von Stumberg, Rui Wang, and Daniel Cremers.
\newblock D3vo: Deep depth, deep pose and deep uncertainty for monocular visual
  odometry.
\newblock In \emph{Proceedings of the IEEE/CVF Conference on Computer Vision
  and Pattern Recognition}, pages 1281--1292, 2020.

\bibitem[{Yin} and {Shi}(2018)]{yin2018geonet}
Zhichao {Yin} and Jianping {Shi}.
\newblock Geonet: Unsupervised learning of dense depth, optical flow and camera
  pose.
\newblock In \emph{2018 IEEE/CVF Conference on Computer Vision and Pattern
  Recognition}, pages 1983--1992, 2018.

\bibitem[Zhan et~al.(2020)Zhan, Weerasekera, Bian, and Reid]{zhan2019visual}
Huangying Zhan, Chamara~Saroj Weerasekera, Jia-Wang Bian, and Ian Reid.
\newblock Visual odometry revisited: What should be learnt?
\newblock In \emph{2020 IEEE International Conference on Robotics and
  Automation (ICRA)}, pages 4203--4210. IEEE, 2020.

\bibitem[Zhang and Tao(2020)]{zhang2020empowering}
Jing Zhang and Dacheng Tao.
\newblock Empowering things with intelligence: a survey of the progress,
  challenges, and opportunities in artificial intelligence of things.
\newblock \emph{IEEE Internet of Things Journal}, 8\penalty0 (10):\penalty0
  7789--7817, 2020.

\bibitem[Zhang et~al.(2021)Zhang, Liu, and Tao]{zhang2018an}
Jingwei Zhang, Tongliang Liu, and Dacheng Tao.
\newblock An optimal transport analysis on generalization in deep learning.
\newblock \emph{IEEE Transactions on Neural Networks and Learning Systems},
  2021.

\bibitem[{Zhou} et~al.(2017){Zhou}, {Brown}, {Snavely}, and
  {Lowe}]{zhou2017unsupervised}
Tinghui {Zhou}, Matthew {Brown}, Noah {Snavely}, and David~G. {Lowe}.
\newblock Unsupervised learning of depth and ego-motion from video.
\newblock In \emph{2017 IEEE Conference on Computer Vision and Pattern
  Recognition (CVPR)}, pages 6612--6619, 2017.

\end{thebibliography}


\end{document}